\def\year{2022}\relax
\documentclass[letterpaper]{article} 
\usepackage{aaai21}  
\usepackage{times}  
\usepackage{helvet} 
\usepackage{courier}  
\usepackage[hyphens]{url}  
\usepackage{graphicx} 
\def\UrlFont{\rm}  
\usepackage{natbib}  
\usepackage{caption} 

\usepackage{physics}
\usepackage{amsmath}
\usepackage{mathtools}
\usepackage{amsfonts}
\usepackage{amsthm}
\usepackage{thm-restate}
\usepackage{ragged2e}
\usepackage{algpseudocode}
\usepackage{algorithm}
\usepackage{delimset}
\usepackage{cleveref}
\usepackage{listings}
\usepackage{subfig}
\usepackage{sgame}
\usepackage{multirow,array}
\usepackage{booktabs}
\usepackage{array, caption, tabularx,  ragged2e,  booktabs}

\usepackage{xcolor}

\newtheorem{definition}{Definition}

\newtheorem{lemma}{Lemma}

\newtheorem{assumption}{Assumption}
\newtheorem{remark}{Remark}

\newcommand{\A}{\mathcal{A}}
\newcommand{\s}{\mathcal{S}}
\newcommand{\M}{\mathcal{M}}
\newcommand{\N}{\mathcal{N}}
\newcommand{\V}{\mathcal{V}}
\newcommand{\E}{\mathcal{E}}
\newcommand{\G}{\mathcal{G}}
\newcommand{\T}{\mathcal{T}}
\newcommand{\I}{\mathcal{I}}
\newcommand{\J}{\mathcal{J}}

\newcommand{\p}{\mathcal{P}}
\newcommand{\X}{\mathcal{X}}
\newcommand{\R}{\mathbb{R}}

\DeclarePairedDelimiterXPP\expect[2]{\mathbb{E}_{#1}}[]{}{\setargs{#2}}%
\NewDocumentCommand{\setargs}{>{\SplitArgument{1}{|}}m}
{\setargsaux#1}
\NewDocumentCommand{\setargsaux}{mm}
{\IfNoValueTF{#2}{#1}{\nonscript\,#1\nonscript\;\delimsize\vert\nonscript\:\allowbreak #2\nonscript\,}}
\DeclarePairedDelimiterXPP\expectaux[3]{\mathbb{E}_{#1}}[]{}{#2\nonscript\:\delimsize\vert\nonscript\:#3}%

\title{Locality Matters: A Scalable Value Decomposition Approach \\ for Cooperative Multi-Agent Reinforcement Learning}

\author{
    Roy Zohar \textsuperscript{\rm 1} ,
    Shie Mannor \textsuperscript{\rm 2},
    Guy Tennenholtz \textsuperscript{\rm 2} \\
}
\affiliations{
    \textsuperscript{\rm 1} Hebrew University of Jerusalem \\
    \textsuperscript{\rm 2} Nvidia Research, Technion Institute of Technology \\
    roy.zohar@mail.huji.ac.il,
    shie@ee.technion.ac.il, 
    guytenn@gmail.com
    }

\begin{document}

\maketitle

\begin{abstract}
Cooperative multi-agent reinforcement learning (MARL) faces significant scalability issues due to state and action spaces that are exponentially large in the number of agents. As environments grow in size, effective credit assignment becomes increasingly harder and often results in infeasible learning times. Still, in many real-world settings, there exist simplified underlying dynamics that can be leveraged for more scalable solutions. In this work, we exploit such locality structures effectively whilst maintaining global cooperation. We propose a novel, value-based multi-agent algorithm called LOMA$Q$, which incorporates local rewards in the Centralized Training Decentralized Execution paradigm. Additionally, we provide a direct reward decomposition method for finding these local rewards when only a global signal is provided. We test our method empirically, showing it scales well compared to other methods, significantly improving performance and convergence speed.
\end{abstract}

\section{Introduction}

The field of Reinforcement Learning (RL) is concerned with an agent taking actions in an environment in order to maximize a cumulative reward. Recent work has witnessed major success in various tasks, including Atari games \cite{atari}, and Go \cite{go}. A popular extension of RL is cooperative multi-agent RL (cooperative MARL), in which a group of agents attempts to interact with an environment together. Research on MARL has gained much attention in recent years, with examples in the Star-Craft multi-agent challenge \cite{starcraft} and traffic control \cite{traffic}.

A common paradigm used in cooperative MARL is Centralized Training Decentralised Execution (CTDE, \citet{ctde}). In this approach, agents are trained simultaneously by a centralized controller. Decentralized policies are then derived from the training process and used for execution. Centralized training can be highly beneficial, granting access to additional global information, which helps agents coordinate their actions. Nevertheless, utilizing such information effectively is a challenging problem for cooperative MARL, due to exponential state and action spaces. As the environment scales, coordination becomes increasingly difficult, rendering centralized training impractical. Still, in many real-world settings, there exist simplified underlying dynamics that can help tackle this problem. 

\begin{figure}[t!]
\centering
\includegraphics[width=0.8\columnwidth]{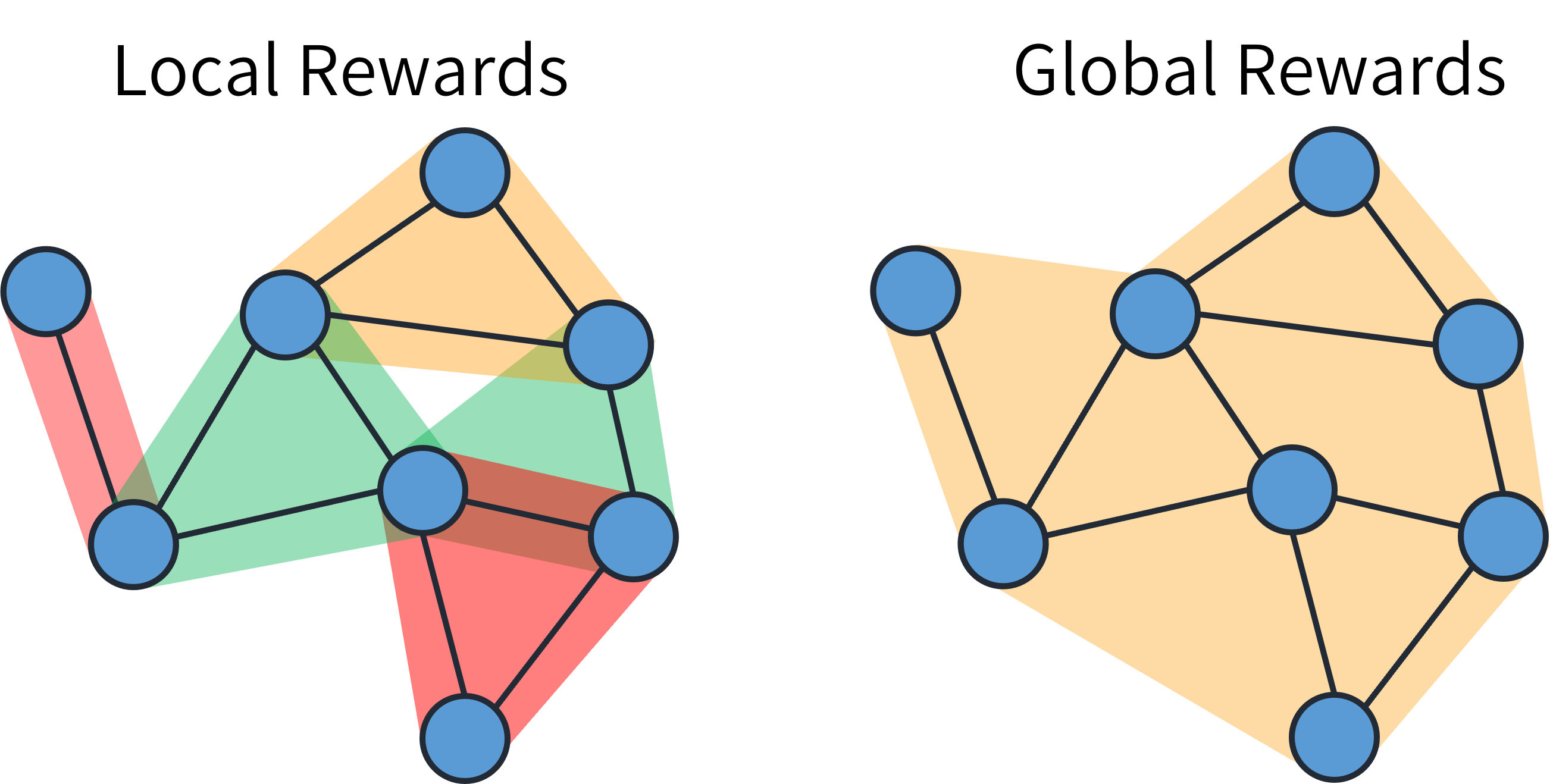}
\caption{A visualization of training MARL for a graph of agents with local rewards vs. with global rewards. The colored regions represent the feedback that the agents exhibit during training}
\label{fig:viz}
\end{figure}

In this paper, we utilize \emph{local rewards}, a principal component of our work. While local rewards are often used in competitive settings (i.e., where every agent attempts to maximize its own local reward), in most cooperative approaches, cooperation is weakly enforced through a shared global reward that all agents aim to maximize \citep{qmix, maddpg}. A visualization of this paradigm is depicted in \Cref{fig:viz}

Local rewards are critical for effective learning in scalable settings. As an example, consider the problem of coaching a large soccer team. If a certain player loses the ball to the other team, punishing that player directly (and possibly neighboring players) with targeted feedback, may be far more effective than punishing the entire team with general feedback. The latter will often leave players confused, believing they should have acted differently.

Despite the effectiveness of local rewards, naively training with local rewards may result in greedy agents that fail to cooperate. Concurrent approaches that aim to exploit local reward structures for our setting often pay a price in terms of cooperation and usually resort to training with global rewards \cite{maddpg}. This is particularly true for the value decomposition approach for cooperative MARL, which has become increasingly popular in recent years \citep{vdn, qmix, qtran, weighted_qmix, qplex}. To the best of our knowledge, there are no value decomposition methods that utilize local rewards effectively. Rather, they rely on the global reward signal for decomposing the joint state-action value function into individual state-action value functions. As we show in our work, such an approach hurts overall performance and convergence speed in large environments. 

In this work, we present a scalable value decomposition method for the cooperative CTDE setting. Our method leverages local agent rewards for improving credit assignment, whilst maintaining a cooperative objective. In addition, we provide a direct decomposition method for finding local rewards when only a global reward is provided. We empirically show that our method is scalable, improving upon state-of-the-art methods for this setting. 

Our contributions are as follows. We define the $Q$-Summation Maximization (QSM) Condition (\Cref{section: summation maximization}), showing its theoretical benefits in a linear bandit setting (\Cref{thm: bandits}). We show that a monotonic decomposition of utilities can be derived to establish the QSM condition (\Cref{section: monotonic decomposition}), and provide a value-based algorithm to enforce it (\Cref{section: LOMA$Q$}). Finally, we construct a reward decomposition method for learning local rewards when a global reward is given (\Cref{subsection: reward decomposition}).

\section{Preliminaries}

We define a multi-agent Markov decision process (MAMDP) as the tuple $\M = (\G, \s, \A, P, r, \gamma)$, where $\G = (\V, \E)$ is an undirected graph of agents, where $\V = [n] = \{1, \hdots, n\}$ and $\E \subseteq \V \times \V$, ${\s = \bigtimes_{i=1}^n \s_i}$ is the global state space, $\A = \bigtimes_{i=1}^n \A_i$ is the global action space, $P: \s \times \s \times \A \mapsto [0,1]$ is the global transition function, $r: \s \times \A \mapsto \R$ is the global reward, and $\gamma \in (0,1)$ is the discount factor.

An agent $i \in \V$ is associated with the underlying graph~$\G$, state $s_i$ and action $a_i$. For a set $B \subseteq \V$ we define $s_B, a_B$ as the subset of agent states and actions in $B$, i.e., ${s_B = (s_i)_{i \in B}}$ and $a_B = (a_i)_{i \in B}$, respectively. At time~$t$, the environment is at state $s = (s_1, \hdots, s_n)$ and the agents take an action $a = (a_1, \hdots, a_n)$, after which the environment returns a reward $r$ and transitions to state $s'$ according to the factored dynamics
$
    P(s'|s, a) 
    = 
    \prod_{i \in \V}
    P_i(s'_i | s_{N(i)}
    , a_i),
$
where here we used $N(i)$ to denote the neighborhood of agent $i$, including $i$, i.e.,
$
    N(i) = \brk[c]*{j \in \V : (i, j) \in \E } \cup \brk[c]*{i}.
$

We define a global Markovian policy $\pi$ as a mapping ${\pi: \s \times \A \mapsto [0,1]}$ such that $\pi(a |s)$ is the probability to choose action $a = (a_0, \hdots, a_n)$ at state $s = (s_0, \hdots s_n)$. We define the value of policy $\pi$ starting at a state $s \in \s$ and taking action $a \in \A$ as
\begin{align*}
    Q^\pi(s,a) 
    = 
    \expect*{\pi}{\sum_{t=0}^\infty \gamma^t r(s(t), a(t)) | s(0) = s, a(0) = a}.
\end{align*}
The value function is then defined by ${v^\pi(s) = \expect*{a \sim \pi(s)}{Q^\pi(s,a)}}$. We define the optimal value and optimal policy by ${v^*(s) = \max_{\pi} v^\pi(s)}$ and ${\pi^* \in \arg\max_{\pi} v^\pi(s)}$, respectively. 

Finally, we denote by $\p$ a partition of $\V = [n]$ (i.e., of agents), such that $\bigcup_{J \in \p} J = \V$ and $\bigcap_{J \in \p} J = \emptyset$. We say that $\p'$ is a refinement of $\p$ if for every $J' \in \p'$ there exists $J \in \p$ such that $J' \subseteq J$\footnote{A refinement partition can be useful when multiple groups of agents concurrently attempt to solve relatively separable tasks.}.

\subsection{Reward Decomposition}
\label{section: preliminaries reward decomposition}

A primary element of MARL is the decomposition of the reward function $r$ over agent states and actions $\brk[c]*{s_i, a_i}_{i \in \V}$. Given some decomposition of rewards $\brk[c]*{r_i : \s \times \A \mapsto \R}_{i \in \V}$, such that $r(s,a) = \sum_{i\in \V} r_i(s,a)$, we define the partial $Q$-function of $\pi$, denoted by $Q_i^\pi(s,a)$ as ${Q_i^\pi(s,a) = \expect*{\pi}{\sum_{t=0}^\infty \gamma^t r_i(s(t), a(t)) | s(0) = s, a(0) = a}}$. It follows that ${Q^\pi(s,a) = \sum_{i \in \V} Q_i^\pi(s,a)}$. Note that such decomposition always exists, e.g., by choosing ${r_1 = r, r_i = 0, i \geq 2}$.

In this work, we consider a reward decomposition for which every agent is dependent only on its local state and action, as defined formally below. We refer the reader to \Cref{section: beyond additive} for a relaxation of this assumption.

\begin{assumption}[\citet{scalable3}]
\label{def:additive1}
We assume that the reward function $r$ is additively decomposable. That is, there exist $\brk[c]*{r_i: \s_i \times \A_i \mapsto \R}_{i \in \V}$ such that $r(s, a) = \sum_{i=1}^n r_i(s_i, a_i)$ for all ${s = (s_1, \hdots, s_n)}$, ${a = (a_1, \hdots, a_n)}$.
\end{assumption}

\section{Value Partitions for MARL}

In this section, we focus on leveraging value-based partitions for credit assignment in MARL. We consider decoupling the problem into smaller problems, each of which can be viewed as a separate, easier estimation problem. Particularly, we generalize ideas from \citet{qmix}, and define a partition-based $Q$-maximization condition. We motivate this condition in a contextual bandit setting, proving it can exponentially improve regret. Then, for the general RL setting, we propose a monotonic decomposition of agent utilities for which our proposed condition holds. We show examples of the latter and prove that monotonic decomposition of utilities is indeed sufficient for partition-based maximization. Our decomposition will prove beneficial in \Cref{section: LOMA$Q$}, as we leverage value partitions to construct a scalable value-based algorithm for MARL.

\subsection{$Q$-Summation Maximization (QSM)}
\label{section: summation maximization}

We begin by defining the $Q$-Summation Maximization condition on which we build upon the rest of this section. The QSM condition states that the $Q$-function can be maximized using a partition of partial maximizers, as defined formally below.

\begin{definition}[QSM Condition]
\label{definition: qsm} Let $\p$ be a partition of $\V$. We say that a MAMDP satisfies the $Q$-Summation Maximisation (QSM) Condition with $\p$, if for every $s \in \s$ and policy~$\pi$
\begin{align*}
    \max_{a}\brk[c]*{\sum_{i=1}^{n}Q_i^\pi(s, a)} =
    \sum_{J \in \p}\brk[r]*{
        \max_{a} \brk[c]*{
            \sum_{i \in J}Q_i^\pi(s, a)
        }
    }
\end{align*}
\end{definition}

We note the two extremes of the QSM Condition. First, every MAMDP satisfies the condition trivially with ${\p= \{\V\}}$. Second, if $\p$ partitions $\V$ into singletons (i.e., $\p= \brk[c]*{\brk[c]*{1}, \brk[c]*{2}, \hdots, \brk[c]*{n}})$, then for every $s \in \s,$
\begin{align*}
    \max_{a}\brk[c]*{\sum_{i=1}^{n}Q_i^\pi(s, a)} =
    \sum^{n}_{i=1}\brk[r]*{
        \max_{a} \brk[c]*{
            Q_i^\pi(s, a)
        }
    }.
\end{align*}

The QSM condition can greatly improve learning efficiency in settings in which the partial $Q$-functions are easier to approximate, effectively decoupling the problem to $\abs{\p}$ simpler problems. We prove this for an instance of the linear bandits problem in the following subsection. Then, in \Cref{section: monotonic decomposition} we discuss a sufficient assumption for which the QSM condition holds.

\begin{algorithm}[t!]
\caption{Multi-OFUL}
\label{alg:multioful}
\begin{algorithmic}[1]
\State {\bf input:} $\alpha, \lambda, \delta > 0$, $\p$ partition of $\V$
\State {\bf init:} $V_{J, a_J} = \lambda I$, $J \in \p, a_J \in \bigtimes_{i \in J} \A_i$. \\
~~~~~~~~~$Y_J = 0$, $J \in \p$.
\For{$t=1,2, \hdots$}
    \State Receive context $x(t)$
    \For{$J \in \p, a_J \in \bigtimes_{i \in J} \A_i$}
        \State $\hat{y}_{a_J}(t) = \brk[a]*{x(t),  V_{J, a_J}^{-1}Y_J}$
        \State $\text{UCB}_{a_J}(t) = \sqrt{\beta_J(t, \delta)}\norm{x(t)}_{ V_{J, a_ J}^{-1}}$
    \EndFor
    \State $a(t) \in \bigtimes_{a_J} \arg\max_{J \in \p} \hat{y}_{a_J}(t) + \alpha \text{UCB}_{a_J}(t)$
    \State Play $a(t)$ and observe $\brk[c]{r_J(t)}$
    \State $V_{J,a_J(t)} = V_{J,a_J(t)} + x(t)x(t)^T, J \in \p$
    \State $Y_J = Y_J + x(t) r_J(t), J \in \p$
\EndFor
\end{algorithmic}
\end{algorithm}

\subsection{QSM in Linear Bandits}
\label{section: linear bandits}

To motivate the QSM condition, we generalize the linear bandit model of \citet{abbasi2011improved}. Specifically, at each round $t$, the environment generates a context ${x(t) \in \X \subseteq \R^d}$ (from a possibly adaptive adversary), where ${\norm{x(t)}_2 \leq S_x}$. The learner must then choose an action ${a(t) \in \A = \bigtimes_{i=1}^n \A_i}$, where $\A_i = [K]$. Given a partition $\p$ of $\V$, the learner then receives $\abs{\p}$ noisy observations $\brk[c]*{r_J(t) = \sum_{i \in J} \brk[a]*{x(t), \theta^*_{i, a_i(t)}} + \eta_J(t)}_{J \in \p}$, where $\brk[c]*{\theta^*_{i, j} \in \R^d: i \in [n], j \in [K]}$ are unknown vectors, $\norm{\theta^*_{i, j}}_2 \leq S_\theta$, and $\brk[c]*{\eta_J(t)}_{J \in \p}$ are independent random variables (for every $t$). We assume $\eta_J(t)$ is conditionally $R_J$-subgaussian random noise, such that 
\begin{align*}
\expect*{}{e^{\lambda \eta_J(t)} | a_J(1), \hdots, a_J(t), \eta_J(1), \eta_J(t-1)} \leq e^{\lambda^2 R_J^2 / 2}.
\end{align*}
We define the regret at time $T$ by
\begin{align*}
    \text{Regret}(T) = \sum_{t=0}^T \sum_{i=1}^n \brk[s]*{\brk[a]*{x(t), \theta^*_{i,a^*_i(t)}} - \brk[a]*{x(t), \theta^*_{i,a_i(t)}}},
\end{align*}
where $a^*(t) \in \arg\max_{a \in \A} \sum_{i=1}^n \brk[a]*{x(t), \theta^*_{i, {a_i}}}$. 

\Cref{alg:multioful} uses the structured partition under which the QSM condition holds. At every iteration of the algorithm, a least square problem is solved for every $J \in \p$, after which an action is chosen according to an upper confidence defined by ${\sqrt{\beta_J(t, \delta)} = \lambda^{1/2}\abs{J}S_\theta + R_{\text{max}} \sqrt{d\log\brk*{\frac{\abs{\p} K^{\abs{J}} (1 + tS_x) / \lambda}{\delta}}}}$. 

Denote $K_\p = \sum_{J \in \p} K^{\abs{J}}$ and ${R_{\text{max}} = \max_{J \in \p} R_J}$. Then, we have the following result.

\begin{restatable}{theorem}{multioful}
\label{thm: bandits}
    Assume $\expect*{}{r_J} \in [-1,1]$ for all $J \in \p$. For all $T \geq 0$, with probability at least $1 - \delta$, the regret of \Cref{alg:multioful} is bounded by
    \begin{align*}
        &\text{Regret}(T) 
        \leq 
        2\sqrt{T}
        \sqrt{
        d
        \log\brk*{\lambda + \frac{T S_x^2}{Kd}}
        K_\p } \times \\
        &\brk*{
        \lambda^{1/2}nS_\theta + R_{\text{max}} \sqrt{d\log\brk*{\frac{\abs{\p} K^n (1 + tS_x) / \lambda}{\delta}}}}.
    \end{align*}
    This leads to, ${\text{Regret}(T) \leq \widetilde{\mathcal{O}}\brk*{dR_{\text{max}}\sqrt{T  K_\p }}}$.
\end{restatable}
The above result achieves regret that is dependent on the maximum subgassuian constant $R_{\text{max}}$ and $\sqrt{K_\p}$. This upper bound is significantly lower than the regret of a naive application of the OFUL algorithm in \citet{abbasi2011improved}. As the latter doesn't assume the QSM condition, it achieves an exponentially larger regret, ${\text{Regret}(T) \leq \widetilde{\mathcal{O}}\brk*{d R_{\text{tot}}\sqrt{T K^n}}}$, where $R_{\text{tot}} = \sum_{J \in \p} R_J$. Indeed, whenever ${\max_{J \in \p} \abs{J} \ll n}$ \Cref{alg:multioful} achieves regret which is exponentially smaller in $K$. Particularly, when $\p = \brk[c]*{\{1\}, \{2\}, \hdots \{n\}}$, we get that ${\text{Regret}(T) \leq \widetilde{\mathcal{O}}\brk*{dR_{\text{max}}\sqrt{T n K}}}$.

\subsection{Monotonic Decomposition of Utilities}
\label{section: monotonic decomposition} 

In \Cref{section: summation maximization} we defined the QSM condition and showed it can significantly improve performance in a linear bandit setting, suggesting its benefits for cooperative MARL. Still, a question arises, when does the QSM condition hold? In this section, we show a sufficient monotonicity assumption under which the QSM condition holds. We formalize this assumption below.
\begin{assumption}[Monotonic Decomposition]
\label{assumption: monotonic utilities}
    We assume there exists a partition $\p$ of $\V$, utility functions $\brk[c]*{U_i^\pi: S_i \times \A_i \mapsto \R}_{i \in \V}$, and partition functions $\brk[c]*{F_J^\pi: \R^n \mapsto \R}_{J \in \p}$ such that for all $J \in \p$, 
    \begin{align*}
        F_J^\pi\brk*{\mathbf{U}(s, a)} &= \sum_{i \in J} Q_i^\pi(s,a), \text{ and} \\
        \nabla_{\mathbf{U}} F_J^\pi &\geq \mathbf{0},
    \end{align*}
    where ${\mathbf{U}(s,a) := (U_1^\pi(s_1, a_1), \hdots, U_n^\pi(s_n, a_n))^T}$.
\end{assumption}
\begin{remark}
    The monotonic decomposition assumption generalizes to trajectory-dependent utilities $U_i^\pi: \T \mapsto \R$, such that
    $F_J^\pi\brk*{\mathbf{U}(\tau)} = \sum_{i \in J} Q_i^\pi(s,a)$, where $s, a$ are the final state and action in the trajectory $\tau$.
\end{remark}
A basic setting for which \Cref{assumption: monotonic utilities} holds is decoupled MAMDPs. Indeed, for any $\M = (\G, \s, \A, P, r, \gamma)$ such that $\E = \emptyset$ and $\M$ is additively decomposable (see \Cref{def:additive1}), we have that \Cref{assumption: monotonic utilities} holds for any partition $\p$. We refer the reader to the appendix for a proof as well as examples of \Cref{assumption: monotonic utilities}. 

\subsection{Monotonic Utilities are Sufficient for QSM}

Next, we show that monotonic utilities (\Cref{assumption: monotonic utilities}) are sufficient for the QSM condition. Additionally, we show that, under \Cref{assumption: monotonic utilities}, local utilities are enough for global $Q$-maximization. This result is closely related to maximization results in previous work \citep{qtran,qmix}. See Appendix for proof.
\begin{restatable}{theorem}{satisfyingqsm}
\label{proposition: satisfyingqsm}
    Suppose \Cref{assumption: monotonic utilities} holds for some partition $\p$. Then the QSM condition (\Cref{definition: qsm}) is satisfied with $\p$. Moreover, for any state $s = (s_1, \hdots, s_n) \in \s$, ${\arg\max_{a \in \A} Q^\pi(s, a) = \bigtimes_{i=1}^n \arg\max_{a_i \in \A_i} U^\pi_i(s_i, a_i)}$.
\end{restatable}

\Cref{assumption: monotonic utilities} is a generalization of the monotonicity assumption of $Q$-mix \citep{qmix}, which holds when ${\p= \{\V\}}$. In contrast, when $\p= \brk[c]*{\brk[c]*{1}, \brk[c]*{2}, \hdots, \brk[c]*{n}}$, each $Q_i$ can be expressed as a function of $\mathbf{U(s, a)}$, and is monotonic w.r.t to its inputs. The following proposition shows that, if \Cref{assumption: monotonic utilities} holds for $\p'$, a refinement of $\p$, then it holds for $\p$ as well. Particularly, this means that if \Cref{assumption: monotonic utilities} holds for any $\p$, then the assumption holds for $\p = \brk[c]*{\V}$.

\begin{restatable}{proposition}{refinement}
\label{proposition: refinement}
    Let $\M = (\G, \s, \A, P, r, \gamma)$, and let $\p, \p'$ be partitions such that $\p'$ is a refinement of $\p$. If $\Cref{assumption: monotonic utilities}$ holds for $\p'$, then it also holds for $\p$.
\end{restatable}

The above proposition suggests a certain trade-off between the refinement of $\p$ and the number of MAMDP's that satisfy \Cref{assumption: monotonic utilities}. \Cref{assumption: monotonic utilities} can thus be viewed as a trade-off between expressibility and speed, as controlled by the refinement of $\p$. 

In the next section, we build upon \Cref{assumption: monotonic utilities} to construct a scalable value-based MARL algorithm that efficiently leverages local value-partitions and local rewards.

\section{Local Multi-Agent $Q$-Learning}
\label{section: LOMA$Q$}

In this section, we describe a value-based approach that leverages the QSM condition using an application of \Cref{assumption: monotonic utilities}. \Cref{alg:local_qmix} provides pseudo-code of our method, which we call LOcal Multi-Agent $Q$-learning (LOMA$Q$). We assume local agent rewards are observable during learning (this assumption will be lifted in \Cref{subsection: reward decomposition}). Instead of approximating the global $Q$-function, LOMA$Q$ builds upon \Cref{assumption: monotonic utilities} to approximate the partition functions $\{F_J\}_{J \in \p}$. 

\begin{algorithm}[t!]
\caption{LOMA$Q$ with local rewards}
\label{alg:local_qmix}
\begin{algorithmic}[1]
\State \textbf{Input:} Partition $\p$ of $\V$, exploration parameter $\epsilon$
\State \textbf{Init:} $F_J(\{\mathbf{U}(s', a')) = 0$, for all $J \in \p$
\For{$t=1,2 \hdots$} 
    \State Take action $a$
    \State Observe $s'$ and local rewards $\brk[c]*{r_J}_{J \in \p}$ 

    \State $a'_{\text{greedy}} \in
    \brk*{\arg\max_{a_i'} U_i(s_i', a_i')}_{i \in \V}$
    \State $a' \gets
    \begin{cases} 
        \text{random action}         &, \text{w.p. } \epsilon \\ 
        a'_{\text{greedy}} &,  \text{w.p. } 1-\epsilon
    \end{cases}$

    \For{$J \in \p$} 
        \State $F_J(\mathbf{U}(s, a)) \overset{\alpha_t}{\gets} r_J(s,a) +
            \gamma F_J(\mathbf{U}(s', a'))$
        \State Project $F_J$ to the set $\brk[c]*{f: \R^n \mapsto \R \text{ s.t. } \nabla f \geq 0}$
    \EndFor
\EndFor
\end{algorithmic}
\end{algorithm}

\Cref{alg:local_qmix} receives as input a partition $\p$ and enforces the monotonicity assumption of $\Cref{assumption: monotonic utilities}$. At every iteration of the algorithm, a greedy action is taken w.r.t. each utility. After an action has been selected, $\brk[c]*{F_J}_{J \in \p}$ are updated using a bellman update for every $J \in \p$. Finally, in line~10, monotonicity is enforced to ensure \Cref{assumption: monotonic utilities} holds. After training is complete, we use the learned utilities $U_i$ for decentralized execution. We note that, due to \Cref{proposition: satisfyingqsm}, choosing the greedy action in line~6 w.r.t. the local utilities is equivalent to acting greedily w.r.t. the global $Q$-function. We refer the reader to the appendix for a discussion regarding the convergence of \Cref{alg:local_qmix}.

\subsection{Practical Implementation of LOMA$Q$}

We implement LOMA$Q$ in a deep $Q$-learning framework \citep{qmix}. Specifically, we approximate $F_J^\pi$ for every $J \in \p$, and $U_i^\pi$ for every ${i \in \V}$ using neural networks with parameters $\theta$. We denote these approximations by $F_J^{\theta}$ and $U_i^{\theta}$, respectively. The outputs of $U_i^\theta$ are forwarded as inputs into $F_J^\theta$, i.e., $F_J^\theta(\{U_i^\theta\}_{i=1}^n)$. 

Given a mini-batch of tuples $(s, a, r, s')$ sampled from a replay memory, we train the neural networks end-to-end by minimizing the loss

\begin{align}
\label{eq: loss}
    L_F(\theta) = \expect*{s, a, s'}{
        \sum_{J \in \p}\brk[r]*{
            y_J - F_J^{\theta}(\{U_i^\theta(s_i, a_i)\}_{i=1}^n)
        }^2
    },
\end{align}
where, 
$
    y_J 
    = 
        \sum_{j \in J}r_j
    +
    \gamma \max_{a'}\brk[c]*{
        F_J^{\theta}(\{U_i^\theta(s_i', a_i')\}_{i=1}^n)
    }.
$

\begin{figure}[t!]
\centering
\includegraphics[width=0.8\columnwidth]{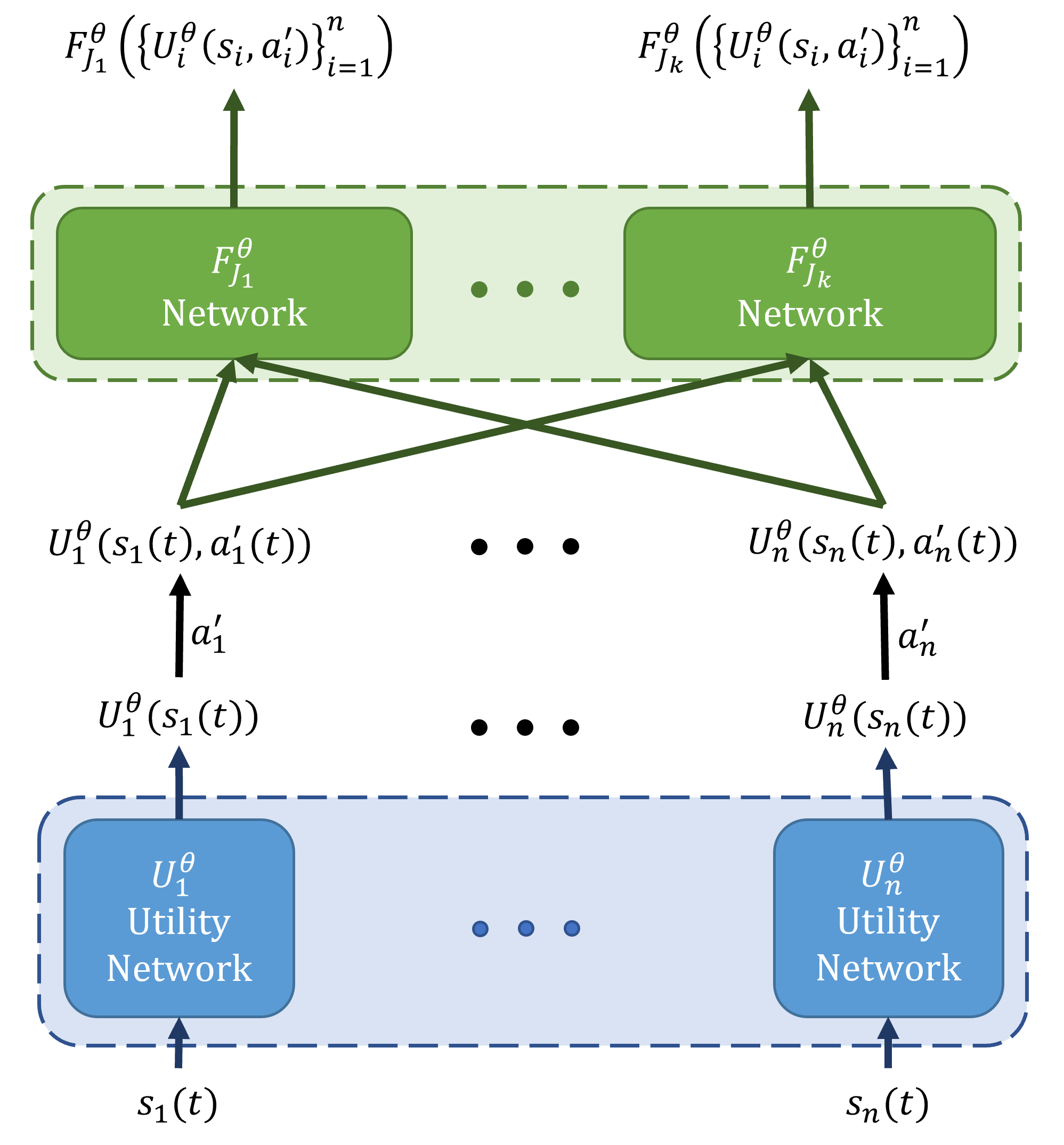}
\caption{The architecture for the LOMA$Q$ network. The agent's states $s_i$ are fed into the utility networks $U_i^\theta$, which are then forwarded as inputs into $F_J^\theta$.}
\label{fig: local qmix architecture}
\end{figure}

\Cref{fig: local qmix architecture} depicts the feed-forward architecture for LOMA$Q$. The local agents states $s_i$ are fed into $U_i^\theta$, which outputs a vector of size $\A_i$, representing the utility of every state-action pair $(s_i, a_i)$. The utilities of the chosen actions $a_i'$ are then forwarded as inputs into $F_J^{\theta}(\{U_i^\theta(s_i, a_i')\}_{i=1}^n)$. Finally, the outputs of $F_J^\theta$ are trained according to $L_F(\theta)$ in \Cref{eq: loss}.

In practice, every agent $i \in \V$ views a trajectory of local states, represented by $\tau_i$. We use recurrent networks for estimating $U_i^\theta$, and fully-connected networks for $F_J^\theta$. We utilize the graph structure for approximating $F_J$, by redirecting $U_i$ into $F_J$ only if there exists a $j \in J$ such that $i \in N(j)$. We refer the reader to the appendix for an exhaustive overview of specific implementation details.

\paragraph{Monotonic Regularization}

To enforce the monotonicity criterion of \Cref{assumption: monotonic utilities}, we implement line~10 of \Cref{alg:local_qmix} by regularizing the loss in \Cref{eq: loss}. We propose two such regularizations; namely, using hard and soft projection regularizers.

For hard regularization, we project all parameters~$\theta$ to be positive through a Relu activation function, i.e., ${\theta \gets \text{Relu}(\theta)}$ for all $\theta$ corresponding to $F_J^\theta$. Alternatively, to allow for softer regularization, we penalize \Cref{eq: loss} by the negative derivatives of $F_J^\theta$ w.r.t. $U_i^\theta$ for every $J \in \p$. That is,
$
    L(\theta) = L_F(\theta)
    +
    \lambda \mathcal{R}_{reg}(\theta),
$
where $\lambda > 0$, and

\begin{align*}
    \mathcal{R}_{reg}(\theta) = \sum_{J \in \p}
      \text{Relu}(-\nabla_{\mathbf{U}} F_J^\theta).
\end{align*}
Here, the regularization parameter $\lambda$ reflects a trade-off between efficiency (due to QSM) and accuracy (whenever \Cref{assumption: monotonic utilities} does not hold exactly).

\subsection{Global Reward}
\label{subsection: reward decomposition}

While LOMA$Q$ relies on observable local rewards for estimating $F_J^\pi$, they may not always be provided. In this section, we propose a new method for decomposing the global reward function into local reward functions, whenever these are not available. 

We assume the global reward signal can be approximately additively decomposed (see \Cref{def:additive1}). We approximate each local reward $r_i(s_i, a_i)$ using a deep neural network with parameters $\phi$. Our prediction for the global reward is then given by $r^\phi_{\text{pred}}(s,a) = \sum_{i=1}^n r_i^\phi(s_i,a_i)$, which is trained to match the global reward signal $r_{\text{global}}$, by minimizing the loss

\begin{align}
\label{eq: reward loss}
    L_r(\phi)
    = \expect*{s,a}{\brk[r]*{
        r^\phi_{\text{pred}}(s, a) 
            - r_{\text{global}}(s,a)
        }^2
    }.
\end{align}

Training $r_i^\phi(s_i, a_i)$ is done in parallel to LOMA$Q$, where $(s, a, r_\textrm{global})$ are sampled from a replay memory. We refer the reader to the appendix for an exhaustive overview and further implementation details.

\subsection{Beyond Additive Decomposition}
\label{section: beyond additive}

In certain settings, \Cref{def:additive1} may be too restrictive, e.g. when interactions between agents are exhibited in the global reward signal. To overcome this, we consider an alternative decomposition of the reward, where every learned reward function can be dependent on a \emph{group} of agents. 

Formally, for any $i \in \V$ we denote by $\I(i)$ the power set of agents in  $\brk[c]*{i} \cup N(i)$. That is,
\begin{align*}
    \I(i) = \brk[c]*{I : I \text{ is in the power set of } \brk[c]*{i} \cup N(i)}.
\end{align*}
Next, for every set $I \in \I(i)$ we define a reward function relating to the agents in $I$, $r_I : \s_I \times \A_I \mapsto \R$. Finally, we define the reward of agent $i \in \V$ by
\begin{align}
\label{eq: power set decomposition}
    r_i(s, a) = \sum_{I \in \I(i)} \frac{1}{|I|}r_I(s_I, a_I),
\end{align}
where here, every reward $r_I$ in the summand is normalized according to the cardinality of $I$. Notice that this decomposition is a generalization of \Cref{def:additive1}. Indeed, \Cref{eq: power set decomposition} coincides with \Cref{def:additive1} whenever $\E = \emptyset$.

The reward decomposition in \Cref{eq: power set decomposition} creates a hierarchy for every agent $i$, as every local reward $r_i$ is comprised of multiple learned reward functions $\brk[c]*{r_I(s_I, a_I)}_{ I \in \I(i)}$ which have less effect on agent $i$ as $|I|$ increases. 

In most cases, the number of local reward functions is exponential in $N(i)$, rendering large decompositions infeasible. Moreover, as $|I|$ increases, the learned rewards become dependent on more agents, reducing their effectiveness (due to normalization in $\abs{I}$). We therefore focus on learning reward functions of small cardinality in $\abs{I}$. We enforce this in practice using a regularization term that is dependent on $\abs{I}$. Specifically, we regularize the loss in \Cref{eq: reward loss} by
\begin{align*}
    \mathcal{R}_{\text{reg}}(\phi)
    = \sum_{I \in \I}  w(|I|) \times |r_I^\phi(s_I, a_I)|,
\end{align*}
where $w(|I|)$ are weights that grow proportionally to $|I|$, penalizing $r_I^\phi(s_I, a_I)$ as $|I|$ increases. This regularization reflects a trade-off between the overall accuracy of the learned rewards and the complexity of the reward decomposition.

\section{Experiments}
\label{section: experiments}

In this section we test the performance of LOMA$Q$ and compare it to previous MARL approaches on two large-scale multi agent tasks.

\subsection{Environments}

We tested our algorithm on two environments, Coupled-Multi-Cart-Pole and Bounded-Cooperative-Navigation. Both environments include minor modifications of the well-known Cart-Pole \cite{cartpole} and Cooperative-Navigation \cite{maddpg} environments. 

\begin{figure}[t!]
\centering
\includegraphics[width=0.8\columnwidth]{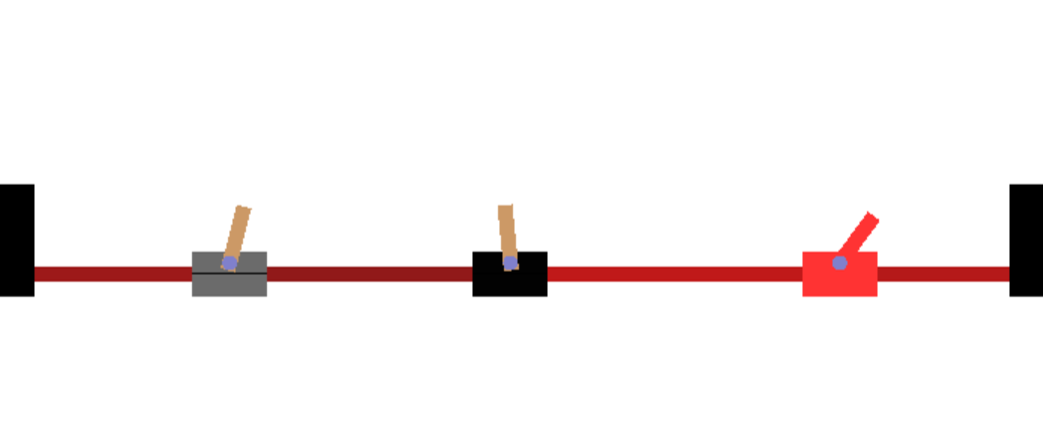}
\caption{The Coupled-Multi-Cart-Pole environment with 3 cartpoles. The right-most cartpole has fallen (marked in red). The global reward for this timestep is +2.}
\label{fig2}
\end{figure}

The Coupled-Multi-Cart-Pole consists of $n$ cartpoles, residing on the $1$d axis. Each cart is viewed as an agent, controlled by applying a force of $\pm1$. Every pair of neighboring carts is connected by a spring. Every cart receives a local reward of $+1$ for every timestep that the pole is upright. The global reward for the environment is the total number of cartpoles that are currently upright. The dependency graph for this environment can be modeled as a line graph, where every cartpole has two neighbors excluding the edges which only have one. \Cref{fig2} depicts this environment for three cartpoles.

\begin{figure}[t!]%
    \centering
    \subfloat[\centering Particles in purple, landmarks in red, regions in gray ]{{\includegraphics[width=0.49\columnwidth]{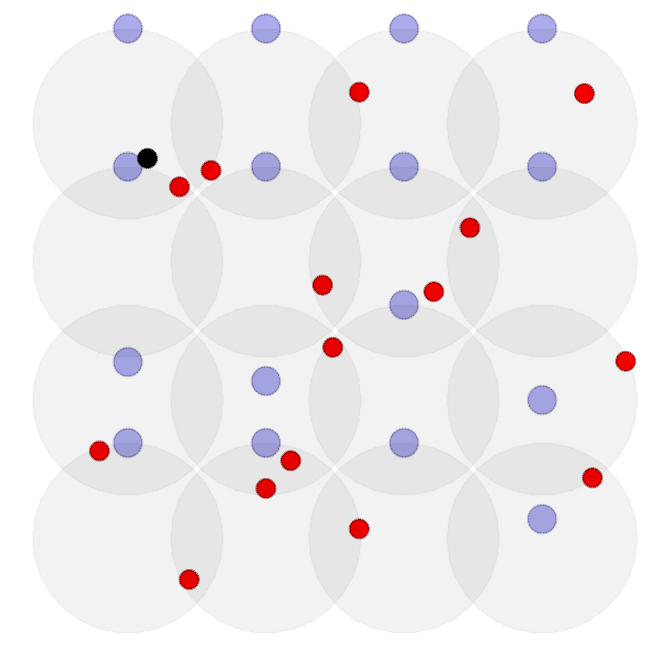} }}%
    \subfloat[\centering Interaction graph based on region overlap]{{\includegraphics[width=0.49\columnwidth]{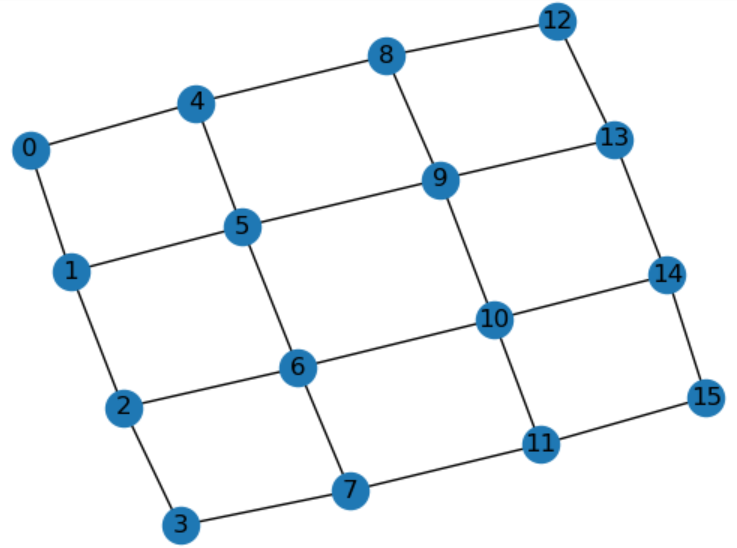} }}%
    \caption{The Bounded-Cooperative-Navigation environment with 16 agents and circular regions.}%
    \label{fig:fig3}%
\end{figure}

The Bounded-Cooperative-Navigation consists of $n$ agents (particles) and $n$ landmarks. Agents must strive to cooperatively cover as many landmarks as possible. In this environment, particles aren't able to move freely in 2d space. Every particle is bound to a fixed distance from its starting position and is thereby restricted to a certain region. This restriction resembles a simplified food-delivery service, where the landmarks represent customers and the agents represent delivery people. Consequently, not all particles interact with each other directly, since direct interactions only occur when two particles are in the same location. This induces a dependency graph for which every two particles are neighbors in the graph if and only if their regions overlap. The environment rewards $+1$ for every landmark that is covered by a particle at a certain timestep. \Cref{fig:fig3} shows a conceptual visualization of the task.

\subsection{Comparative Experiments}

\subsubsection{Scalability} We tested LOMA$Q$ on both cooperative environments with $n=15$ agents in two setups; namely, with and without access to a local reward signal. We denote these by LOMA$Q$ and LOMA$Q$+RD, respectively. In both cases, we used the refined partition $\p= \brk[c]*{\brk[c]*{1}, \brk[c]*{2}, \hdots, \brk[c]*{n}}$. 

We compared LOMA$Q$ to a wide range of contemporary cooperative methods. In addition, we compared LOMA$Q$ to two versions of IQL \cite{IQL}, trained with environment local rewards and global rewards, which we denote by IQL-local and IQL, respectively. 

\begin{figure*}[t!]
\centering
\includegraphics[width=2\columnwidth]{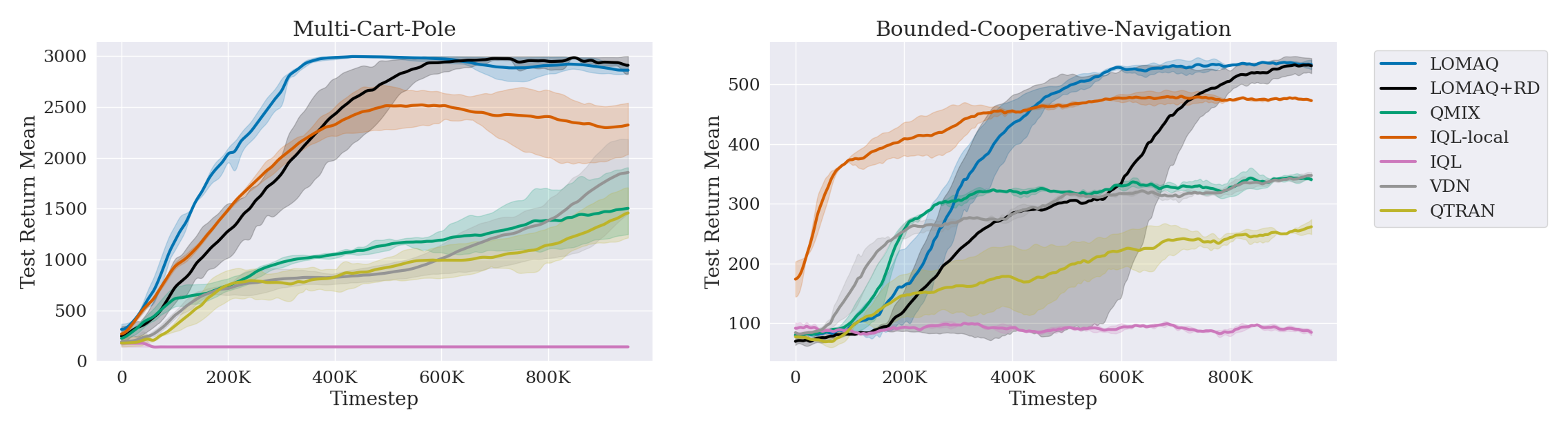}
\caption{Test returns for the Coupled-Multi-Cart-Pole environment and for the Bounded-Cooperative-Navigation environment.}
\label{fig:mutli_cart}
\end{figure*}

\Cref{fig:mutli_cart} depicts the results of the Coupled-Multi-Cart-Pole and Bounded-Cooperative-Navigation environments. It is evident that both versions of LOMA$Q$ significantly outperform all of the compared methods in both performance and convergence speed. We note that LOMA$Q$+RD converges to LOMA$Q$'s policy, with a slight delay due to the time taken to learn the reward decomposition. 

In both environments various cooperative methods exhibit slow learning compared to LOMA$Q$, due to the use of global rewards. Additionally, while IQL-local does learn quickly (primarily due to the use of local rewards), it converged to a sub-optimal solution. This occurs as IQL acts greedily w.r.t its local rewards. In contrast, LOMA$Q$ incentivizes cooperation, enabling both fast convergence as well as improved performance. 

\subsubsection{Reward Decomposition} 
\begin{figure}[t!]
\centering
\includegraphics[width=\columnwidth]{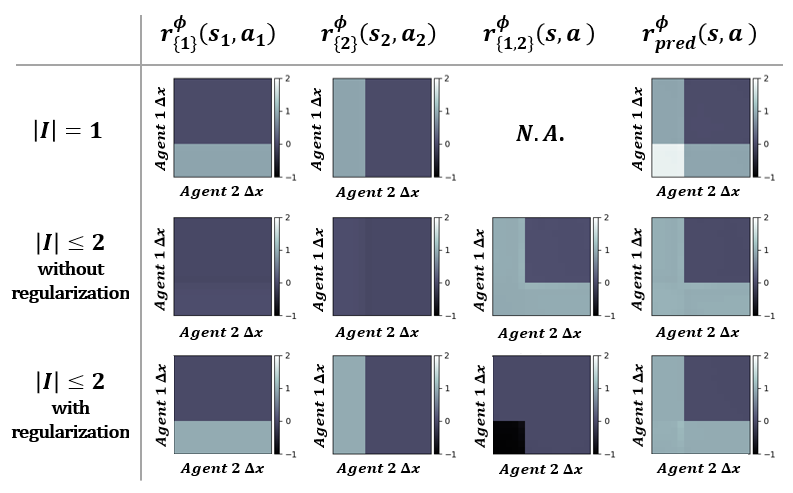}
\caption{Visualization of learned reward functions $r_I^\phi$ for different decompositions in Bounded-Cooperative-Navigation for $n=2$ agents and a single landmark. We plot the learned reward functions as a function of Agent 1 and Agent 2's distance from the landmark, denoted by $\Delta x$. The first row assumes $|I| \leq 1$, the second row assumes $|I| \leq 2$ with no regularization, and the third row adds regularization.}
\label{fig:viz_decompose}
\end{figure}

We visualize multiple reward decompositions for Bounded-Cooperative-Navigation. We run our decomposition method with a global reward signal, for $n=2$ agents and a single landmark. If both agents are on the landmark at the same time, the global reward remains 1. We plot the learned reward functions as a function of Agent~1 and Agent~2's distance from the landmark, which we denote by $\Delta x$. These results are depicted in \Cref{fig:viz_decompose}.  

The first row in \Cref{fig:viz_decompose} assumes a decomposition according to \Cref{def:additive1}. \Cref{def:additive1} does not hold for this setup, since the reward function is dependent on both agents when they share a landmark. The approximated reward is overly optimistic and wrongly rewards $+2$ when the landmark is shared. The second row approximates a decomposition that allows $|I| \leq 2$ with no regularization $\lambda =0$. In this case, the global reward is approximated correctly, and the local reward functions $r^\phi_{\brk[c]*{i}} = 0$. The third row visualizes a decomposition with regularization $w(|I|=2)=1, \lambda=0.0001$. In this case, the local reward functions $r^\phi_{\brk[c]*{i}}$ convey information for each agent $i$, and $r^\phi_{\brk[c]*{1, 2}}$ conveys information regarding their joint dynamic. 

\section{Related Work}

\paragraph{Graph Based MARL.} The underlying structure of the team of agents in the environment can often be modeled using a graph topology. \citet{DGN} propose DGN - a MARL algorithm based on the graph convolutional network (GCN) architecture which assumes centralized execution and homogeneous agents. \citet{graphmix} propose GraphMIX for CTDE, that uses global rewards for learning. \citet{scalable3} propose Scalable Actor-Critic - An Actor-Critic approach for the discrete space case which utilizes a dependency graph with theoretical guarantees. 

\paragraph{Cooperative MARL.} Our approach enhances the popular value decomposition family \citep{qtran, qplex, weighted_qmix}, which consider a cooperative multi-agent problem in which each agent observes its own state and action history. \citet{vdn} propose VDN for decomposing the value function into a sum of utility functions. \citet{qmix} offer $Q$-mix which generalizes this concept, by decomposing the value function into a monotonic function of individual utility functions. All of these approaches implicitly measure the impact of every agent on the observed global reward, whereas we propose to combine this line of work with an explicit approach for credit assignment using local rewards. 

\paragraph{Credit Assignment.} Various approaches have attempted to tackle the credit assignment problem. A common approach for credit assignment is by estimating the individual $Q$ functions $Q_i$ directly, which are often substantially simpler and significantly easier to learn than $Q$ \cite{scalable3, orig_drQ, drsarsa, hra, proof_drQ}. Our work extends this line of work for value-based CTDE, and focuses on reward decompositions that expedite learning alongside global cooperation. 

\paragraph{Reward Decomposition.} Multiple works recognize the benefits of local rewards and attempt to learn them in settings where only a global reward signal is provided. $RD^2$ \cite{RD2} learns a reward decomposition with minimally-dependent features for factored-state MDP setting. Our method can be seen as an extension of $RD^2$ for MARL, where the action is also factored. 

\paragraph{Large Action Spaces.} Finally, our work is related to work on large and combinatorial action spaces. From action elimination \citep{zahavylearn}, to action embeddings \citep{tennenholtz2019natural,chandak2019learning}, through action redundancy \citep{baram2021action}, our work can be viewed as an additional method for reducing the effective dimensionality of the problem.

\section{Conclusion and Future Work}

In this work we tackled the credit assignment problem of cooperative MARL through local, partition based value functions. We used the QSM condition and a monotonic decomposition of utilities to construct a value-based approach, effectively reducing the problem to simpler ones. We showed that local rewards are highly beneficial, both when provided as well as learned implicitly from a global reward. These greatly improved overall performance and convergence speed, suggesting that local structures can be efficiently used to improve MARL algorithms.

In this work we have assumed that an underlying, static dependency graph $\G$ is provided during training. In many cases, these assumptions are limiting. We look to further generalize our method by learning such dynamic dependencies between agents through interaction with the environment. In addition, our work has assumed that $\Cref{assumption: monotonic utilities}$ holds for some partition $\p$ and local reward decomposition $\{r_i\}$. We look to generalize our algorithm to automatically identify effective decompositions.

\newpage
\bibliography{bibliography.bib}

\onecolumn
\section*{Appendix A: Monotonic Decomposition}

This section further analyzes the properties of the monotonic decomposition that we introduced in \Cref{section: monotonic decomposition}. We introduce a few additional results that (proofs can be found in Appendix D), discuss the limitations of \Cref{assumption: monotonic utilities} with a bandit example, and formally justify the approximation of truncated $F_J$ mentioned in \Cref{section: LOMA$Q$}. 

\subsection*{Additional Results}

First, the following proposition claims that under \Cref{assumption: monotonic utilities}, the global $Q$ function is also monotonic w.r.t the utility functions. This is a useful and important result that we use in multiple proofs.

\begin{restatable}{proposition}{globalmonotonicity}[Monotonicity of the global $Q$ function]
\label{proposition: global monotonicity} 
Suppose \Cref{assumption: monotonic utilities} holds for some partition $\p$. Then the global $Q$ function $Q^\pi(s, a)$ can be written as a function of utilities $\mathbf{U(s, a)}$ and is monotonic w.r.t every utility
\end{restatable}

In \Cref{section: monotonic decomposition} we claimed that \Cref{assumption: monotonic utilities} holds for decoupled MAMDPs. The following proposition formalizes this claim.

\begin{restatable}{proposition}{decoupled}
\label{proposition: decoupled}
    Let $\M = (\G, \s, \A, P, r, \gamma)$ such that $\E = \emptyset$ and $\M$ is additively decomposable by \Cref{def:additive1}. Then \Cref{assumption: monotonic utilities} holds for any partition $\p$
\end{restatable}

In this work, we have considered reward decompositions that are additively decomposable, such that $r(s, a) = \sum_{i=1}^n r_i(s_i, a_i)$. This assumption is limiting in many cases, and can be generalized in many ways, for instance with $\beta_2$ as done in \cite{scalable4}.

In practice, even if local rewards depend on a small group of agents, they still expedite LOMA$Q$ substantially. The following proposition, shows that the refinement of $\p$ can be deepened by modifying the reward decomposition. Note that the mentioned reward decompositions are dependent on the \emph{global} state-action $r_i(s, a)$.

\begin{restatable}{proposition}{modified}
    Let $\M = (\G, \s, \A, P, r, \gamma)$ such that $\Cref{assumption: monotonic utilities}$ holds from some partition $\p$ and reward decomposition $\brk[c]*{r_i(s,a)}_{i=1}^n$. Then, for any refinement $\p'$ of $\p$ there exists a reward decomposition $\brk[c]*{r_i'(s,a)}_{i=1}^n$ such that $\sum_{i=1}^n r_i'(s,a) = \sum_{i=1}^n r_i(s,a)$ and $\Cref{assumption: monotonic utilities}$ holds with $\p'$.
\end{restatable}

In the extreme case where $\p=\{\V\}$ and $\p'= \brk[c]*{\brk[c]*{1}, \brk[c]*{2}, \hdots, \brk[c]*{n}}$, the proposed reward decomposition assumes the global reward function for every agent: $r_i'(s,a) = \frac{1}{n}\sum_{j \in \V} r_j(s, a)$. 

\subsection*{Limitations of Monotonic Utilities}

\begin{figure}[t!]
\centering
\includegraphics[width=0.6\columnwidth]{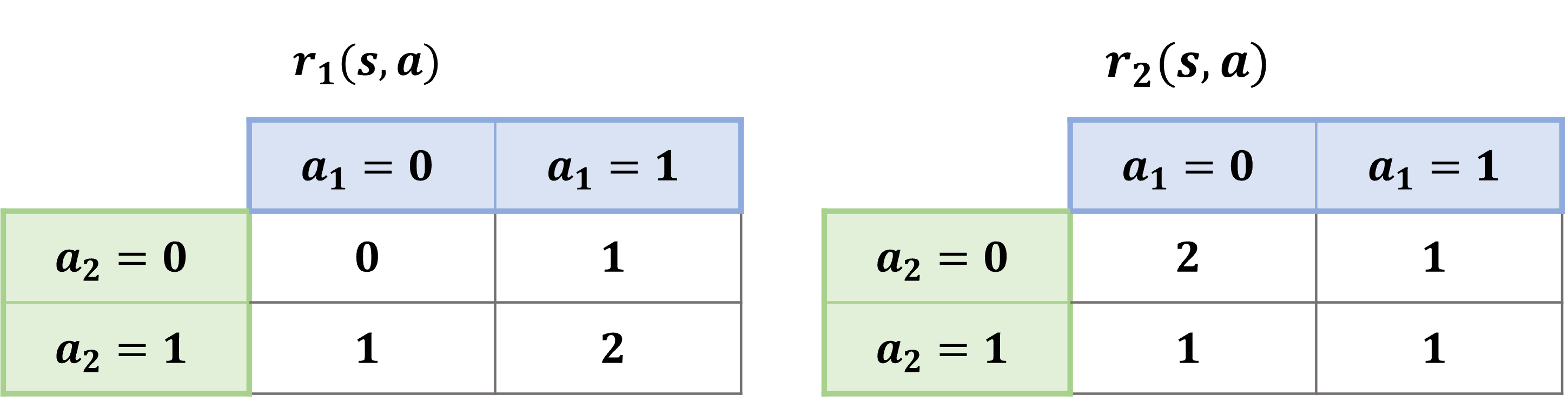}
\caption{Payoff matrices with 2 local reward functions. $Q_1$ and $Q_2$ are monotonic w.r.t to different utility functions.}
\label{fig:payoff1}
\end{figure}

In this section, we formalize an example that outlines some limitations of \Cref{assumption: monotonic utilities}. \Cref{fig:payoff1} depicts a  bandit setting example (with $\gamma = 0$) for which \Cref{assumption: monotonic utilities} only holds for certain partitions. Note that the reward decomposition in \Cref{fig:payoff1} does not satisfy \Cref{def:additive1} purposely, since bandit settings that satisfy \Cref{def:additive1}, satisfy \Cref{assumption: monotonic utilities} as an immediate corollary of \Cref{proposition: decoupled}.

For the payoff matrices in \Cref{fig:payoff1}, \Cref{assumption: monotonic utilities} holds for $\p = \brk[c]*{\brk[c]*{1,2}}$ by setting $U_i^\pi(s_i, a_i) = a_i$, and $ Q^\pi(U_1^\pi, U_2^\pi) = F^\pi_{\brk[c]*{\brk[c]*{1,2}}} = 2 + \max\brk[c]*{0, U_1^\pi + U_2^\pi - 1}$ which is monotonic. 

Moreover, there exist $F_{\{1\}}^\pi$, $F_{\{2\}}^\pi$ that are monotonic w.r.t \emph{different} utility functions, e.g.,

\begin{align*}
    U_i^\pi(s_i, a_i) = \begin{cases}
      a_i & \text{for} F^\pi_{\{1\}}\\
      1 - a_i & \text{for} F^\pi_{\{2\}}\\
    \end{cases}
\end{align*}

such that

\begin{align*}
    {F^\pi_{\{1\}} = U_1^\pi + U_2^\pi}
\end{align*}
\begin{align*}
    {F^\pi_{\{2\}} = 1 + \max\brk[c]*{0, U_1^\pi + U_2^\pi - 1}}.
\end{align*}

Which are both monotonic. Still, it can be shown that \Cref{assumption: monotonic utilities} does not hold for $\p = \{\{1\}, \{2\}\}$. To see this, assume by contradiction that \Cref{assumption: monotonic utilities} holds for $\p = \{\{1\}, \{2\}\}$. Let $s \in S$ denote some global state, and $a_x, a_y \in \A$ denote global actions such that $a_x = (a_1, a_2) = (0, 0)$, and $a_y = (a_1, a_2) = (1, 0)$.

By definition $F^\pi_{\{1\}} = Q_1(s, a)$. Therefore, $F^\pi_{\{1\}}(s, a_x) = 0$, and $F^\pi_{\{1\}}(s, a_y) = 1$. Note that $U^\pi_2(s_2, (a_y)_2) = U^\pi_2(s_2, 0) = U^\pi_2(s_2, (a_x)_2)$, since $U^\pi_2$ is not dependent on $a_1$. Therefore, we can write:

\begin{align*}
    F^\pi_{\{1\}}(s, a_x) 
    = F^\pi_{\{1\}}(U^\pi_1(s_1, 1), U^\pi_2(s_2, 0)) = 0
\end{align*}
\begin{align*}
    F^\pi_{\{1\}}(s, a_y) 
    = F^\pi_{\{1\}}(U^\pi_1(s_1, 0), U^\pi_2(s_2, 0)) = 1
\end{align*}

Fixing $U^\pi_2(s_2, 0)$, $F^\pi_{\{1\}}$ is monotonically non-decreasing w.r.t to $U^\pi_1$ due to \Cref{assumption: monotonic utilities}. Since $F^\pi_{\{1\}}(s, a_y) > F^\pi_{\{1\}}(s, a_x)$, then $U^\pi_1(s_1, 0) > U^\pi_1(s_1, 1)$. 

Conversely, since $F^\pi_{\{2\}}(s, a_x) > F^\pi_{\{2\}}(s, a_y)$ and since $F^\pi_{\{2\}}$ is also monotonically non-decreasing w.r.t to $U^\pi_1$, we get that $U^\pi_1(s_1, 0) < U^\pi_1(s_1, 1)$. This of course in contradiction to $U^\pi_1(s_1, 0) > U^\pi_1(s_1, 1)$. 

\subsection*{Truncated Value Decomposition}
\label{subsection: truncated}

In this subsection, we expand upon our approximated decomposition scheme, which leverages the specific topological properties of the underlying agent graph $\G$. While \Cref{assumption: monotonic utilities} defines $F_J$ as functions of \emph{global} states and actions, by using a graph-dependent additive reward decomposition (see \Cref{section: preliminaries reward decomposition}), this assumption can be simplified to \emph{local} states and actions, characterized by neighboring agents.

First, we define the $\kappa$-hop neighborhood of agent $i \in \V$ $\N_i^\kappa$, as the set of all agents that are at most $\kappa$ edges from $i$:  $\N_i^\kappa = \brk[c]*{j \in \V | \text{dist}(i, j) \leq \kappa}$. Note that for $\kappa = 1$ we get our original neighborhood definition $N(i)$ from \Cref{section: preliminaries reward decomposition}.

Theoretically, every $F_J$ is somewhat dependent on the entire global action space $(s, a)$. However, agents that are distant from agent $i$ in $\G$, have a negligible effect on agent $i$. This result is especially important in terms of credit assignment - by truncating distant agents, credit assignment becomes substantially easier and increases the effectiveness of each local reward.

\cite{scalable3} study this effect, and define truncated $Q$-functions $\tilde{Q}^\pi_i(s_{N^{\kappa}_i}, a_{N^{\kappa}_i})$ that are only dependent on a local neighborhood $N^{\kappa}_i$. We extend this result for our case be defining truncated $F_J^\pi$, such that $\tilde{F}_J^\pi = \sum_{i \in J}\tilde{Q}^\pi_i(s_{N^{\kappa}_i}, a_{N^{\kappa}_i})$. The following proposition describes the accuracy of this approximation. 

\begin{restatable}{proposition}{truncated}[Truncated Decomposition]
\label{proposition: truncated}
    If the reward function is additively decomposable, and the exponential decay property \cite{scalable3} holds with $(c, \rho)$, then for every $J \in \p$, and $(s,a) \in \s \times \A$,
    \begin{align*}
        |F^\pi_J - \tilde{F}_J^\pi| \leq |J|c\rho^{\kappa + 1}
    \end{align*}
\end{restatable}

Appendix B describes the full use of the truncated $\tilde{F}_J^\pi$ in LOMA$Q$. 

\newpage
\section*{Appendix B: Experiments}

This section depicts further details regarding our experimentation. We give further explanations regarding LOMA$Q$ and our reward decomposition method, depict the full algorithms and the used hyper-parameters for both methods, and provide a couple of further ablations for LOMA$Q$. 

All experiments in the main paper were run with 3 randomly selected seeds. Every 10,000 time-steps, we tested all algorithms in a decentralized manner on 20 episodes with $\epsilon=0$ and recorded the test return mean. This metric is suitable since some methods implement global mixing layers that defy decentralized execution. We averaged the test return mean for every seed, and plotted the result. We also added the minimal and maximal value for every test in a translucent color. All experiments were run on various Linux machines with varying architectures (including both CPU and GPU GeForce RTX 2080 Ti) and memory constraints. We refer the reader to the README.md file in the code that describes the full procedure for replicating our results. We did a coarse independent sweep for every hyper-parameter. 

\subsection*{Additional Implementation Details}

\subsubsection{LOMA$Q$} Similar to previous work \cite{atari}, we store two sets of parameters to stabilize the training of LOMA$Q$: $\theta$ for the current network parameters, and $\theta^-$, as target parameters that are used for future estimates. The target parameters are updated at a slower time scale. Additionally, we implement parameter sharing between networks. If agents are homogeneous, we share parameters between $U_i^\theta$.

\citet{qmix} propose using \emph{hyper-networks} that depend on the global state for generating non-negative weights for $Q$. LOMA$Q$ also allows the use of hyper-networks for every $F_J^\theta$, however since (1) the number of hyper-networks can be very large ($O(|J|)$) and (2) for large settings only a small portion of the global state is relevant, our implementation uses hard regularization that is independent of the global state. We leave testing LOMA$Q$ with hyper-networks and global "sub-states" as future work. 

As shown in \Cref{proposition: truncated}, it is often useful to approximate $F_J^\theta$ with a local neighborhood $\N_i^\kappa$. In practice, we redirect $U_j^\theta$ into $F_J^\theta$, if and only if there exist $i, j$ such that $i \in J$ and $j \in \N_i^\kappa$, where $\kappa$ is a hyper-parameter that controls the accuracy of approximation $F_J^\theta$. We have also implemented this redirection using a GCN (Graph Convolutional Network), where the number of convolution layers is $\kappa$, which yielded similar results. This concept is similar to the mixing layer in Graph-mix \cite{graphmix}.

\subsubsection{Reward Decomposition} We chose a non-recurring architecture for approximating $r_I^\phi$, since each decomposition should be independent of other time-steps. If the reward is independent of agent id, we assume parameter sharing between classes of $r_I^\phi$ with equal $|I|$.

We note that the approximated $r_I^\phi$ are only used when $r^\phi_{\text{pred}}$ achieves approximate convergence to $r_{\text{global}}$. That is, we only use $r_I^\phi$ if ${\abs{r^\phi_{\text{pred}}(s, a)  - r_{\text{global}}} \leq \Delta}$ where $\Delta$ is a tolerance hyper-parameter. If the inequality doesn't hold, we disregard $r_I^\phi$. 

We have also implemented a classification variant of our decomposition method that assumes that for every $I \in \I$, and $(s, a) \in \s \times \A$, $r_I(s_I, a_I) \in K_I$, where $K_I$ is some small group of possible reward values. This method is exponential in $n$, but converged very well for small $n$'s. Nevertheless, all experiments use the regression variant of our decomposition method, depicted in \Cref{subsection: reward decomposition}.

\subsection*{Full Algorithms}

\subsubsection*{LOMA$Q$}

\Cref{alg:local_qmix_full} includes the full training scheme for LOMA$Q$, outlined in \Cref{alg:local_qmix}. The full algorithm consists of 4 main parts (1) Storing trajectories in the replay buffer (2) Training the reward decomposition if we are running LOMA$Q$-RD (3) Updating our estimation for the $Q$-values and (4) Enforcing monotonicity. The algorithm references \Cref{alg:reward_decomposition_train} and \Cref{alg:reward_decomposition_infer}, that describe our reward decomposition method and are presented in the next section.

\begin{algorithm*}[t!]
\caption{Full algorithm for training LOMA$Q$ / LOMA$Q$+RD}
\label{alg:local_qmix_full}
\begin{algorithmic}[1]
\Require MAMDP $\M$, partition $\p$, hyperparameters $\kappa$, $\epsilon$, $\alpha$, $L$
\State Initialize replay memory D
\State Initialize $[F_J^\theta]$, $[U_i^\theta]$ with random parameters $\theta$
\State Redirect $[U_i^\theta]$ outputs into $F_J^\theta$ only if there exists a $j \in J$ such that $i \in \N_j^\kappa$ according to $\G$
\State Initialize target parameters $\theta^- = \theta$
\For{$\text{episode}=1\hdots$} 
    \State \textit{// Sample and Store a trajectory}
    \State Observe initial state $s(0)$
    \For{$t=0, \hdots$} 
        \State $a'_{\text{greedy}} \in
        \brk*{\arg\max_{a_i'} U_i(\tau_i'(t), a_i')}_{i \in \V}$
        \State $a'(t) \gets
        \begin{cases} 
            \text{random action}         &, \text{w.p. } \epsilon \\ 
            a'_{\text{greedy}} &,  \text{w.p. } 1-\epsilon
        \end{cases}$
        \State Take action $a'(t)$ and retrieve next observation $s(t+1)$ and reward $r(t)$
        
        \If{LOMA$Q$+RD and $r(t)$ is global}
            \State $r(t) \gets$ \Cref{alg:reward_decomposition_infer}
        \EndIf
        \State Store transition $(\tau(t), a(t), r(t), \tau(t+1))$ in D
    \EndFor
        
    \\
    \State \textit{// Train Decomposer}
    \If{LOMA$Q$+RD}
        \Cref{alg:reward_decomposition_train}
    \EndIf

    \\
    \State \textit{// Train Agents}

    \State Sample a random mini-batch $B$ of transitions $(\tau, a, r, \tau')$ from D

    \State Update $\theta$ by minimizing the loss using learning rate $\alpha$ whilst enforcing monotonicity
    \begin{align*}
        L_F(\theta) = \sum_{(\tau, a, r, \tau') \in B}\brk[r]*{
            \sum_{J \in \p}\brk[r]*{
                y_J - F_J^{\theta}(\{U_i^\theta(s_i, a_i)\}_{i=1}^n)
            }^2
        },
    \end{align*}
    \begin{align*}
        y_J 
        = 
            \sum_{j \in J}r_j
        +
        \gamma \brk[r]*{
            F_J^{\theta^-}\brk[r]*{\brk[c]*{
                \max_{a'_i}\brk[c]*{
                    U_i^{\theta^-}(\tau_i', a'_i)
                }
            }_{i=1}^n}
        }
    \end{align*}
    \State \textit{// Enforce Monotonicity}
    \If{Hard regularization}
        \State Enforce Monotonicity by $\theta \gets \text{Relu}(\theta)$
    \Else 
        \State $L_F(\theta) \gets L_F(\theta) + \lambda \mathcal{R}_{reg}(\theta)$, where $ \mathcal{R}_{reg}(\theta) = \sum_{J \in \p}
        \text{Relu}(-\nabla_{\mathbf{U}} F_J^\theta)$
    \EndIf

    \\
    \State Anneal $\epsilon$
    \State Update target network parameters $\theta^- \gets \theta$ with period $L$
\EndFor
\end{algorithmic}
\end{algorithm*}

\subsubsection{Reward Decomposition} \Cref{alg:reward_decomposition_train} and \Cref{alg:reward_decomposition_infer} depict our reward decomposition method in full. \Cref{alg:reward_decomposition_train} describes the training procedure of every $r_I^\phi$ using a global reward signal, and \Cref{alg:reward_decomposition_infer} describes the prediction stage, where $r_I^\phi(s_I, a_I)$ are inferred and translated into local agent rewards as seen in \Cref{section: beyond additive}.

\begin{algorithm}[t!]
\caption{Training Reward Decomposition}
\label{alg:reward_decomposition_train}
\begin{algorithmic}[1]
\Require MAMDP $\M$, Replay memory D, hyperparameters $\I$, $\alpha_2$, $\lambda$
\If{not initialized}
    \State Initialize $[r_I]_{I \in \I}$ with random parameters $\phi$
\EndIf    
    
\State Sample a random mini-batch $B$ of transitions $(s, a, r_{\text{global}})$ from D
\State $r^\phi_{\text{pred}}(s,a) \gets \sum_{I \in \I} r_I^\phi(s_I, a_I)$
\State Update $\theta$ by minimizing the loss with learning rate $\alpha_2$
\begin{align*}
       L_r(\theta)
        = \sum_{(s, a, r_{\text{global}}) \in B}{\brk[r]*{
            r^\phi_{\text{pred}}(s, a) 
                - r_{\text{global}(s,a)}
            }^2
        } 
        + \lambda \mathcal{R}_{\text{reg}}(\phi)
\end{align*}
\begin{align*}
    \mathcal{R}_{\text{reg}}(\phi)
        = \sum_{(s, a, r_{\text{global}}) \in B}{\brk[r]*{
            \sum_{I \in \I} w(|I|) \times |r_I^\phi(s_I, a_I)|
        }}
\end{align*}
\end{algorithmic}
\end{algorithm}

\begin{algorithm}[t!]
\caption{Inferring Reward Decomposition}
\label{alg:reward_decomposition_infer}
\begin{algorithmic}[1]
\Require $(s, a, r_{\text{global}})$, Trained Networks $\brk[c]*{r_I(\theta)}_{I \in \I}$, hyper-parameter $\Delta$

\State $r_{\I} \gets \brk[c]*{
    r_{I}(s_{I}, a_{I} ; \theta)
    }_{I \in \I}$
\State $r_{\text{pred}} \gets 
    \sum_{r_{I} \in r_{\I}}
    r_{\I}
$

\If{$|r_{\text{pred}} - r_{\text{global}}| \geq \Delta$}
    \State\Return Decomposition has failed 
\EndIf

\State $ r_{i} \gets \brk[s]*{
    \sum_{r_{I} \in r_{\I}}
    \frac{1}{|I|}r_{I}
}_{i=1}^n$
\State\Return $r_{i}$
\end{algorithmic}
\end{algorithm}

\subsection*{Hyper-Parameters} 
\Cref{table:hyper} depicts the hyper-parameters used in LOMA$Q$ and \Cref{table:hyper2} depicts additional hyper-parameters used in our reward decomposition method for LOMA$Q$-RD. 

\begin{table*}[t!]
\centering
\begin{tabular}{p{0.2\columnwidth}p{0.3\columnwidth}p{0.4\columnwidth}}
  \toprule
    \textbf{Hyper-parameter}             & \textbf{Values}                  & \textbf{Description}       \\ 
  \midrule
         $\p$    & 
         $\p= \brk[c]*{\brk[c]*{1}, \brk[c]*{2}, \hdots, \brk[c]*{n}}$                   & 
        Partition used in LOMA$Q$
    \\
        $\epsilon$    & 
        Linear anneal from 1.0 to 0.05 for 100K timesteps                   & 
        Parameter used for $\epsilon$-greedy action selection
    \\
        $\gamma$    & 
        0.99                   & 
        Used in Bellman updates
    \\
        L Target update interval    & 
        50 episodes                  & 
        Frequency of switching between $\theta$ and $\theta^-$
    \\
        batch size    & 
        50                   & 
        Batch of trajectories sampled from replay memory
    \\
        learning rate $\alpha$    & 
        0.0005                   & 
        Used with RMSProp optimizer. Also used $\alpha_{\text{RMS}}=0.99$, $\epsilon_{\text{RMS}}=0.00001$
    \\
        $\kappa$    & 
        1                   & 
        Radius of neighborhood taken for estimating $F_J$. Equal to $N(i)$ in this case.
    \\
        Monotonicity Method    & 
        Hard                   & 
        We used hard regularization, $\theta \gets \text{Relu}(\theta)$
    \\
        Layers in $U_i^\theta$    & 
        Linear, Relu, GRU, Linear                   & 
        Input shape is obs size, 64 for hidden dim in GRU, output shape is N\underline{o} actions
    \\
        Parameter Sharing in $U_i^\theta$    & 
        all                   & 
        Both environments include homogeneous agents
    \\
        Layers in $F_J^\theta$    & 
        Linear, elu, Linear, elu, Linear                   & 
        Input shape for $F_J^\theta$ is $|\bigcup_{j \in J}N(j)|$, 32 for hidden dim, output shape is 1
    \\
        Parameter Sharing in $F_J^\theta$    & 
        Multi-Cart-Pole: ${J \in \brk[c]*{\brk[c]*{2}, \brk[c]*{3}, \hdots, \brk[c]*{13}}}$, 
        Bounded-Cooperative-Navigation:  ${J \in \brk[c]*{\brk[c]*{3}, \brk[c]*{4}, \hdots, \brk[c]*{12}}}$, 
        & 
        Since both environments exhibit symmetry, we share parameters between inner $F_J^\theta$ since mixing should be similar between them. 
    \\
  \bottomrule
\end{tabular} 
\caption{Hyper-parameters used for LOMA$Q$ in both environments}
\label{table:hyper} 
\end{table*}

\begin{table*}[t!]
\centering
\begin{tabular}{p{0.2\columnwidth}p{0.3\columnwidth}p{0.4\columnwidth}}
  \toprule
    \textbf{Hyper-parameter}             & \textbf{Values}                  & \textbf{Description}       \\ 
  \midrule
         $\I$    & 
         $\I= \brk[c]*{\brk[c]*{1}, \brk[c]*{2}, \hdots, \brk[c]*{n}}$                   & 
        We only decomposed the reward according to \Cref{def:additive1} in LOMA$Q$+RD. In Bounded-Cooperative-Navigation, we allowed every agent to observe how many agents are currently with it on the landmark, therefore making the reward additively decomposable by \Cref{def:additive1}. This also effectively means $\lambda = 0$
    \\
        $\Delta$    & 
        n * 0.1 = 1.5                   & 
        Parameter used for disregarding unsuccessful decompositions
    \\
        batch size    & 
        5                   & 
        Batch of trajectories sampled from replay memory for reward decomposition
    \\
        learning rate $\alpha_2$    & 
        0.01                   & 
        Used with Adam optimizer with default settings. 
    \\
        Layers in $r_I^\phi$    & 
        Linear, Relu, Linear, Leaky Relu, Linear, Tanh, Linear                   & 
        Input shape is obs size * $|I|$, 64 for first hidden dim, 128 for second hidden dim, output shape is 1
    \\
        Parameter Sharing in $r_I^\phi$    & 
        all                   & 
        Both environments include homogeneous agents, and $|I| \leq 1$
    \\
  \bottomrule
\end{tabular} 
\caption{Hyper-parameters used for reward decomposition}
\label{table:hyper2} 
\end{table*}

\subsection*{Additional Experiments}

\subsubsection{Representability}

\begin{figure}[t!]
\centering
\includegraphics[width=0.6\columnwidth]{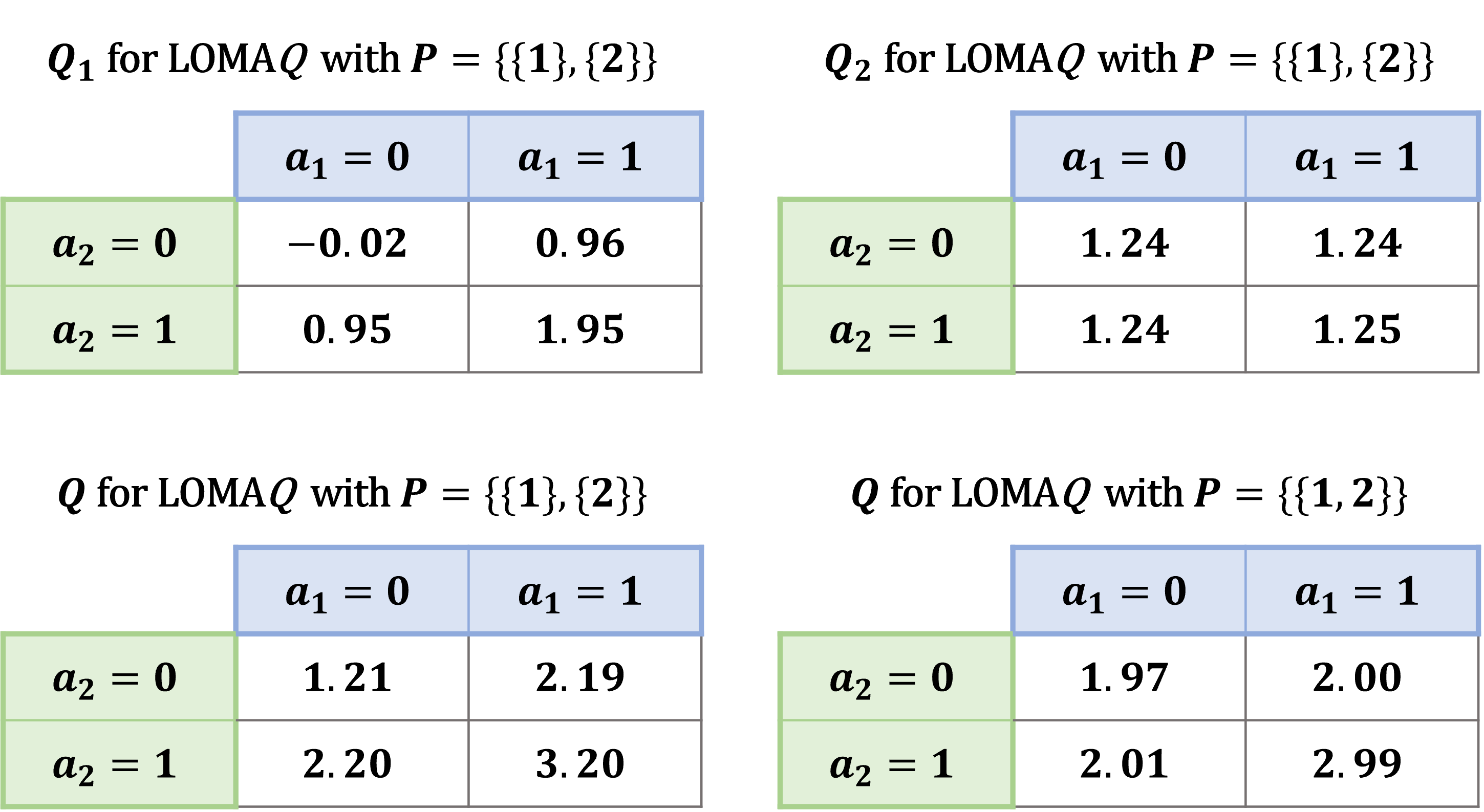}
\caption{Learned $Q$ values for $P_1 = \{\{1, 2\}\}$, but not for $P_2 = \{\{1\}, \{2\}\}$}
\label{fig:payoff_results}
\end{figure}

We evaluate the representability of LOMA$Q$ on the matrix game described in \Cref{fig:payoff1} for different partitions. We set $\gamma=0$ for simplicity, such that $Q_i(s,a) = r_i(s,a)$, and $\epsilon=1$ for testing representability of the entire state-action space. The results can be seen in \Cref{fig:payoff_results}. As expected LOMA$Q$ converged for $P = \{\{1, 2\}\}$ but not for $P = \{\{1\}, \{2\}\}$. 

Nevertheless, we note that even when $P = \{\{1\}, \{2\}\}$, although the values of  $Q_2^\theta$ are wrong, the values of $Q_1^\theta$ are approximately correct, and the global $Q^\theta$ resembles the correct $Q$. We believe this is due to the fact that $Q_1$ induces larger TD-errors, that have a larger effect on the utilities.

\subsubsection{Truncated Approximation $\kappa$} 
\begin{figure}[t!]
\centering
\includegraphics[width=0.6\columnwidth]{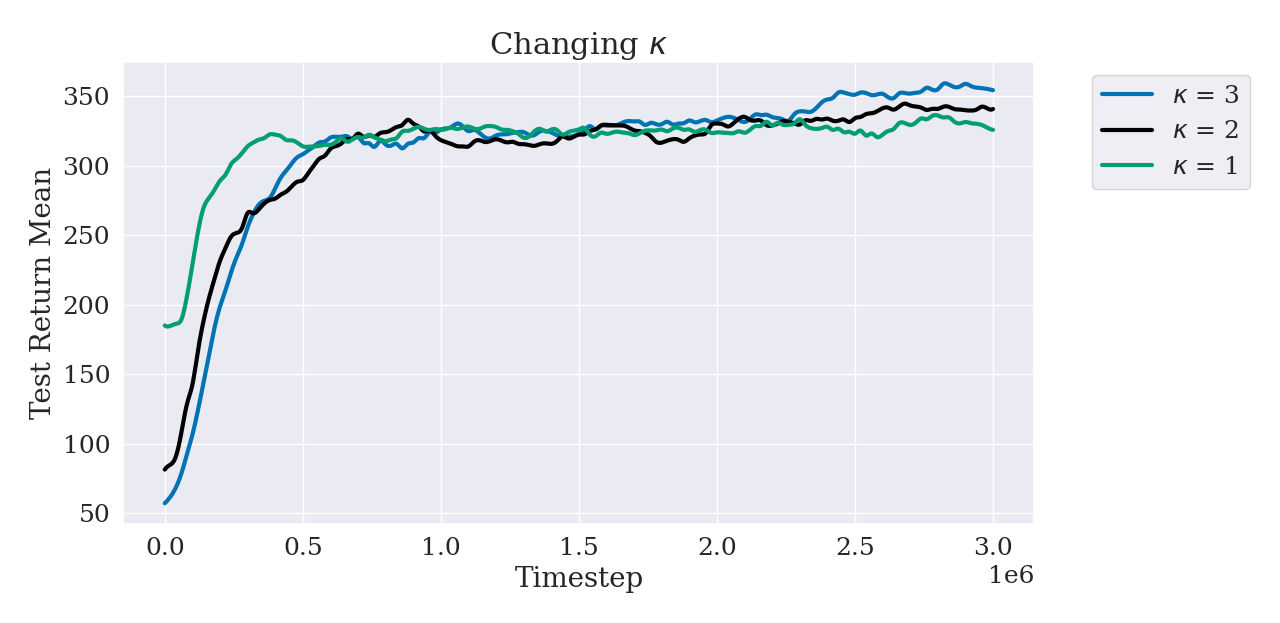}
\caption{learned $Q$ values changing $\kappa$}
\label{fig:changing_k}
\end{figure}

We test our algorithm by changing $\kappa$, that control the accuracy of approximation of every $F_J^\theta$ on Multi-Cart-Pole. These results can be seen in \Cref{fig:changing_k}. As can be seen in the figure, convergence is slightly slower as $\kappa$ grows, and the converged policy is slightly better. We believe that for environments that are coupled in a stronger manner, this effect will be more dominant in both convergence speed and performance. 

\clearpage

\section*{Appendix C: Discussion}

This section presents a few further discussions regarding our work. We further compare our work to existing methods, present the differences between our methods for enforcing monotonicity in LOMA$Q$, and discuss the convergence of LOMA$Q$.  

\subsection*{Relation to other works}
\label{subsection: other works}

\subsubsection{$Q$-mix \cite{qmix}} LOMA$Q$ can be seen as a generalization of $Q$-mix that incorporates local rewards. The training process of LOMA$Q$ can be described as training $|\p|$ instances of $Q$-mix, that are trained on shared utility functions. For $\p = \{\V\}$, LOMA$Q$'s architecture is equivalent to $Q$-mix, since $F_J = F_{\V} = Q$. 

Our soft regularization monotonicity constraint can replace the hyper-network scheme proposed in \cite{qmix}, and offers an alternative, soft approach for satisfying the IGM condition. 

\subsubsection{VDN \cite{vdn}} 

Secondly, our reward decomposition method resembles the VDN architecture for the supervised case. This choice is very suitable, since both the VDN decomposition and our reward decomposition are additive. 

\subsection*{Comparison of monotonicity methods}
\label{subsection: monotonicity comparisons}

In this section, we note 2 important differences between our hard and soft projection regularizers. For brevity, we denote these variations LOMA$Q$-h and LOMA$Q$-s. 

First, LOMA$Q$-h \textbf{\textit{must}} train and enforce monotonicity according to the \textbf{\textit{same partition}} $\p$. Note however that LOMA$Q$-s permits us to approximate $F_J$ where $J \in \p$, whilst enforcing \Cref{assumption: monotonic utilities} according to $\p'$, where $\p$ is only required to be a \textbf{\textit{refinement}} of $\p'$, by enforcing monotonicity for sums of $F_J$ where $J \in \p$. 

This feature can be beneficial if one wishes to estimate $Q$-values according to a refined $\p$ (i.e: explainability \cite{proof_drQ}), for problems that only satisfy $\Cref{assumption: monotonic utilities}$ with a coarse $\p'$. Note that in this case, LOMA$Q$ will probably not see a large increase in training speed, since $\p'$ is coarse.

Another important difference is that while optimizing LOMA$Q$-s, our optimizer can explore non-monotonic parameterizations $\theta$ since we enforce monotonicity softly, whereas LOMA$Q$-h enforces a hard monotonicity constraint that allows the only optimization of monotonic parameterizations.

\subsection*{Convergence of LOMA$Q$}
\label{subsection: convergence}

In this section, we discuss details regarding the convergence of LOMA$Q$. \cite{proof_drQ} formalize dr$Q$, and offer convergence guarantees:

\begin{itemize}
    \item dr$Q$ is guaranteed to converge to the global optimal $\sum_{c} Q_c \mapsto Q^*$. 
    \item For every $c$, $Q_c \mapsto Q_c^*$
\end{itemize}

Note that in Algorithm 1 from \cite{proof_drQ}, the action $a' = a(t+1)$ is chosen w.r.t to the global $Q$ function: $a' = \arg\max_{a}\sum_{i=1}^n Q_{c_J}(s, a)$. This is a vital element for convergence to the global optimum. Similarly, the corresponding line in LOMA$Q$ (\Cref{alg:local_qmix}, Line 6), maximizes over the utility functions, and due to \Cref{proposition: satisfyingqsm} this is equivalent to maximizing the global $Q$ function directly.

Therefore, LOMA$Q$ differs from dr$Q$ solely due to the monotonicity enforcement at every update. Due to these enforcements, the theoretical convergence results do not carry over directly and require a stronger argument. We leave the rigorous proof of convergence for LOMA$Q$ as future work. 

\newpage
\section*{Appendix D: Missing Proofs}

This section includes all of the missing proofs from the paper. This section is divided into 2 main parts, (1) proofs regarding monotonic decomposition and (2) proofs regarding linear bandits. 

\subsection{Proofs for Monotonic Decomposition}

\globalmonotonicity*
\begin{proof}

Since $Q^\pi(s,a) =  \sum_{i=1}^n Q_i^\pi(s,a) = \sum_{J \in \p}F_J(\mathbf{U}(s,a))$, we can denote $Q^\pi(\mathbf{U}(s,a)) = Q^\pi(s,a)$. 

Secondly, we have that $Q^\pi$ is monotonic with respect to every $U_i^\pi \in \mathbf{U}$, as

\begin{align*}
    \forall i \in \V:
    \pdv{Q^\pi}{U^\pi_i} 
    = \pdv{}{U^\pi_i}\brk[r]*{
        \sum_{J \in \p}\sum_{j \in \J}Q^\pi_j
    } 
    = \pdv{}{U_i^\pi}\brk[r]*{
        \sum_{J \in \p}F_J^\pi
    } 
    = \sum_{J \in \p}\brk[r]*{
        \pdv{F_J^\pi}{U_i^\pi} 
    } \geq 0
\end{align*}
\end{proof}

\satisfyingqsm*
\begin{proof}

We will start by proving that the QSM condition holds. Let $s \in \s$ be some global state, and $\pi$ be some global Markovian policy. For every $i \in \V$, let $u^*_{i}$ denote the maximal value for every $U_i^\pi$ i.e. $u^*_{i} = \max_{a_i} \{U_i^\pi(s_i, a_i)\}$.

Due to the monotonicity of $\brk[c]*{F_J^\pi}_{J \in \p}$ and $Q^\pi$ w.r.t to their inputs $\mathbf{U(s, a)}$ (\Cref{proposition: global monotonicity}), we can maximize each of these functions by maximizing $\mathbf{U(s, a)}$. That is:

\begin{align}
\label{equation: almost_igm}
    \max_a \left\{Q^\pi(\mathbf{U}(s,a))\right\}
    = Q^\pi(\{
        \max_{a_i}\{U_i^\pi(s_i, a_i)\}
    \}_{i \in \V}),
\end{align}and that: \begin{align*}
    \max_a \left\{F_J^\pi(\mathbf{U}(s,a))\right\}
    = F_J^\pi(\{
        \max_{a_i}\{U_i^\pi(s_i, a_i)\}
    \}_{i \in \V})
\end{align*}

So overall we get that

\begin{align*}
    \max_{a}\brk[c]*{\sum_{i=1}^{N}Q_i^\pi(s, a)} 
    & = \max_a \left\{Q^\pi(s, a)\right\} \\
    & = \max_a \left\{Q^\pi(\mathbf{U}(s,a))\right\} \\
    & = Q^\pi(\{
        \max_{a_i}\{U_i^\pi(s_i, a_i)\}
    \}_{i \in \V}) \\
    & = Q^\pi(\{u^*_i\}_{i \in \V}) \\
    & = \sum_{J \in \p} \brk[r]*{
        \sum_{j \in J} Q_j^\pi(\{u^*_i\}_{i \in \V})
    } \\
    & = \sum_{J \in \p}\brk[r]*{
        \max_{a} \brk[c]*{
            \sum_{j \in J}Q_j^\pi(s, a)
        }
    }.
\end{align*}

Next, we show that ${\arg\max_{a \in \A} Q^\pi(s, a) = \bigtimes_{i=1}^n \arg\max_{a_i \in \A_i} U^\pi_i(s_i, a_i)}$. This condition closely resembles the IGM Condition from \citep{qtran, qmix}. This is straightforward due to \Cref{equation: almost_igm}, completing the proof.
\end{proof}

\refinement*
\begin{proof}

Let $\p'$ be a refinement of $\p$. That is, for every $J' \in \p'$, there exists a $J \in \p$ such that $J' \subseteq J$. Let $\brk[c]*{F_{J'}^\pi: \R^n \mapsto \R}_{J' \in \p'}$, $\brk[c]*{U_i^\pi: S_i \times \A_i \mapsto \R}_{i \in \V}$ be functions that satisfy \Cref{assumption: monotonic utilities} for $\p'$.

We define new functions $\brk[c]*{F_J^\pi: \R^n \mapsto \R}_{J \in \p}$ as follows:

\begin{align*}
    \forall J \in \p:
    F_J^\pi(\mathbf{U}(s,a)) :=
    \sum_{J' \in \p' | J' \in J} F_{J'}^\pi(\mathbf{U}(s,a))
\end{align*}

Note that \Cref{assumption: monotonic utilities} is satisfied for $\p$, with the newly defined $\brk[c]*{F_J^\pi}_{J \in \p}$ and the original utilities $\brk[c]*{U_i^\pi}_{i \in \V}$, since

\begin{align*}
    \forall J \in \p:
    \nabla_{\mathbf{U}} F_J^\pi
    = \nabla_{\mathbf{U}}\brk[r]*{\sum_{J' \in \p' | J' \in J} F_{J'}^\pi}
    = \sum_{J' \in \p' | J' \in J} \nabla_{\mathbf{U}} F_{J'}^\pi
    \geq \mathbf{0}.
\end{align*}

\end{proof}

\decoupled*
\begin{proof}

Due to \Cref{proposition: refinement}, it suffices to show that \Cref{assumption: monotonic utilities} holds for $\p= \brk[c]*{\brk[c]*{1}, \brk[c]*{2}, \hdots, \brk[c]*{n}}$. Since $\E = \emptyset$, the global transition can be written as:

\begin{align*}
    P(s'|s, a) 
    = 
    \prod_{i \in \V}
    P_i(s'_i | s_{N(i)}
    , a_i),
    = \prod_{i \in \V}
    P_i(s'_i | s_i, a_i),
\end{align*}

This allows us to decouple the $\M$ into $|n|$ completely independent MDPs. For every $i \in \V$, $M_i$ is defined with state space $S_i$, action space $A_i$, reward function $r_i(s_i, a_i)$ and transtion $P_i(s'_i | s_i, a_i)$.

Due to this decoupling, we can write:

\begin{align*}
    \forall i \in \V:
    Q_i^\pi(s,a) = Q_i^\pi(s_i, a_i)
\end{align*}

We can then define $\brk[c]*{F_J^\pi}_{J \in \p}$, $\brk[c]*{U_i^\pi}_{i \in \V}$ for the partition $\p= \brk[c]*{\brk[c]*{1}, \brk[c]*{2}, \hdots, \brk[c]*{n}}$:

\begin{align*}
    \forall i \in \V:
    U_i^\pi(s_i, a_i) := Q_i^\pi(s_i, a_i)
\end{align*}

\begin{align*}
    \forall J \in \p:
    F_J^\pi(\mathbf{U}(s,a)) =
    F_{\{i\}}^\pi(\mathbf{U}(s,a)) =
    Q_i^\pi(s,a) =
    Q_i^\pi(s_i, a_i) =
    U_i^\pi(s_i, a_i)
\end{align*}

And therefore for every $J \in \p$, $\nabla_{\mathbf{U}} F_J^\pi = \nabla_{\mathbf{U}} U_i^\pi(s_i, a_i) = \mathbf{e_i} \geq \mathbf{0}$

\end{proof}

\modified*
\begin{proof}

Assume that \Cref{assumption: monotonic utilities} holds for partition $\p$ with $\brk[c]*{F_J^\pi}_{J \in \p}$, $\brk[c]*{U_i^\pi}_{i \in \V}$, and let $\p'$ be a refinement of $\p$. We denote $J'(i)$ as the parent set $J' \in \p'$ of $i \in \V$. Due to $\p'$ being a partition of $\V$, $J'(i)$ exists and is singular. Similarly, for every $J' \in \p'$ we denote $A(J')$ as the set in $\p$ that containts $J'$. Since $\p'$ is a refinement of $\p$ $A(J')$ exists and is singular.

We define a new reward decomposition $\brk[c]*{r_i'(s,a)}_{i=1}^n$ for $\p'$ as follows:

\begin{align*}
    \forall i \in \V: 
    r_i'(s,a) = \frac{1}{|A(J'(i))|}\sum_{j \in A(J'(i))} r_j(s, a)
\end{align*}

Note that for this new decomposition

\begin{align*}
    \sum_{i=1}^n r_i'(s,a)
    & = \sum_{i=1}^n \brk[r]*{
        \frac{1}{|A(J'(i))|}\sum_{j \in A(J'(i))} r_j(s, a)
    }\\
    & = \sum_{J' \in \p'} \brk[r]*{
         \frac{|J'|}{|A(J')|}\sum_{j \in A(J')} r_j(s, a)
    }\\
    & = \sum_{J \in \p} \brk[r]*{
        \sum_{J' \in \p' | J' \subseteq J} \brk[r]*{
            \frac{|J'|}{|J|}\sum_{j \in J} r_j(s, a)
        }
    }\\
    & = \sum_{J \in \p} \brk[r]*{
        \frac{1}{|J|} \brk[r]*{
             \sum_{J' \in \p' | J' \subseteq J} |J'|
        }\brk[r]*{
            \sum_{j \in J} r_j(s, a)
        }
    }\\
    & = \sum_{J \in \p} \brk[r]*{
        \sum_{j \in J} r_j(s, a)
    } \\
    & = \sum_{i=1}^n r_i(s, a)
\end{align*}

For this new decomposition, also note that

\begin{align*}
    {Q'}_i^\pi(s,a) = 
    & = \expect*{\pi}{
        \sum_{t=0}^\infty \gamma^t r'_i(s(t), a(t)) | s(0) = s, a(0) = a
    } \\
    & = \expect*{\pi}{
        \sum_{t=0}^\infty \gamma^t \brk[r]*{
            \frac{1}{|A(J'(i))|}\sum_{j \in A(J'(i))} r_j(s, a)
        }
        | s(0) = s, a(0) = a
    } \\
    & = \frac{1}{|A(J'(i))|}\sum_{j \in A(J'(i))} \brk[r]*{
        \expect*{\pi}{
            \sum_{t=0}^\infty \gamma^t r_j(s(t), a(t)) | s(0) = s, a(0) = a
        }
    } \\
    & = \frac{1}{|A(J'(i))|}\sum_{j \in A(J'(i))} Q_j^\pi(s,a)
\end{align*}

We can now define $\brk[c]*{{F'}_{J'}^\pi}_{J' \in \p'}$, $\brk[c]*{{U'}_i^\pi}_{i \in \V}$ for the new reward decomposition. We set the utilities to be identical to the utilities from before ${U'}_i^\pi = U_i^\pi$. Therefore:

\begin{align*}
    \forall J' \in \p':
    {F'}_{J'}^\pi(\mathbf{U}(s,a))
    & = \sum_{i \in J'} {Q'}_i^\pi(s,a) \\
    & = \sum_{i \in J'} \frac{1}{|A(J'(i))|}\sum_{j \in A(J'(i))} Q_j^\pi(s,a) \\
    & = \sum_{i \in J'} \frac{1}{|A(J')|}\sum_{j \in A(J')} Q_j^\pi(s,a) \\ 
    & = \frac{|J'|}{|A(J')|}\sum_{j \in A(J')} Q_j^\pi(s,a) \\
    & = \frac{|J'|}{|A(J')|} F_{A(J')}
\end{align*}

Which therefore means:

\begin{align*}
    \nabla_{\mathbf{U}} {F'}_{J'}^\pi 
    = \nabla_{\mathbf{U}} \brk[r]*{
        \frac{|J'|}{|A(J')|} F_{A(J')}
    } 
    = \frac{|J'|}{|A(J')|} \nabla_{\mathbf{U}} F_{A(J')} \geq \mathbf{0}
\end{align*}

Therefore satisfying \Cref{assumption: monotonic utilities} with partition $\p'$ and reward decomposition $\brk[c]*{r_i'(s,a)}_{i=1}^n$.

\end{proof}

\truncated*
\begin{proof}
    
Assume that the exponential decay property holds with $(c, \rho)$, \cite{scalable3} showed that the truncated $Q$-functions prove to be a good approximation to the partial $Q_i^\pi(s,a)$:

\begin{align*}
    \forall (s, a) \in S \times A:
    |
    Q^\pi_i(s, a)
    -
    \tilde{Q}^\pi_i(s_{N^{\kappa}_i}, a_{N^{\kappa}_i})
    |
    \leq 
    c\rho^{\kappa + 1}
\end{align*}

This results immediately carries over for our $F_J^\pi$. We define the truncated $F_J^\pi$ as $\tilde{F}_J^\pi = \sum_{j \in J}\tilde{Q}^\pi_i(s_{N^{\kappa}_i}, a_{N^{\kappa}_i})$. Therefore for every 
$(s, a) \in S \times A$: 

\begin{align*}
        |F^\pi_J(s, a) - \tilde{F}_J^\pi|
        & = \left|
            \sum_{j \in J}Q^\pi_i(s, a)-
            \sum_{j \in J}\tilde{Q}^\pi_i(s_{N^{\kappa}_i}, a_{N^{\kappa}_i})
        \right| \\
        & = \left|\sum_{j \in J}\brk[r]*{
            Q^\pi_i(s, a) - \tilde{Q}^\pi_i(s_{N^{\kappa}_i}, a_{N^{\kappa}_i})
        }\right| \\
        & \leq \sum_{j \in J}\left|
            Q^\pi_i(s, a) - \tilde{Q}^\pi_i(s_{N^{\kappa}_i}, a_{N^{\kappa}_i})
        \right|\\
        & \leq |J|c\rho^{\kappa + 1}
\end{align*}

\end{proof}

\subsection*{Proofs for Linear Bandits}

\multioful*
\begin{proof}
    Denote $\theta^*_{a_J} = \sum_{i \in J}\theta^*_{i, a_i}$ and let $\hat{\theta}_{a_J}(t) = V_{J, a_J}(t)^{-1}Y_J(t)$ be the least squares estimator at time step $t$, where $
         V_{J, a_J}(t)
         =
         \lambda I
         +
         \sum_{k=1}^{t-1}
         \mathbf{1}_{\brk[c]*{a_J(k) = a_J}}x(k)x(k)^T$.
    Let ${\sqrt{\beta_J(t, \delta)} = \lambda^{1/2}\abs{J}S_\theta + R_{\text{max}} \sqrt{d\log\brk*{\frac{\abs{\p} K^{\abs{J}} (1 + tS_x) / \lambda}{\delta}}}}$ and
    \begin{align*}
        \mathcal{C}_{t,a_J}
        =
        \brk[c]*{\theta \in \R^d : \norm{\theta_{a_J} - \hat{\theta}_{a_J}(t)}_{V_{J, a_J}} \leq \sqrt{\beta_J(t)}}.
    \end{align*}
    
    Define the good events $\G_J = \brk[c]*{ \theta^*_{a_J} \in \mathcal{C}_{t,a_J}, \forall t \geq 0, a_J \in \bigtimes_{i \in J} \A_i }$. By \Cref{lemma: confidence ellipsoid}, $P(G_J) \geq 1 - \frac{\delta}{\abs{\p}}$. By the union bound, $P\brk*{\bigcup_{J \in \p} \G_J} \geq 1 - \delta$. Denote by $I_k(J, a_J) = \min\brk[c]*{t : \sum_{i=1}^t \mathbf{1}_{\brk[c]*{a_J(k) = a_J}} = k}$ the $k^{th}$ time action $a_J$ was chosen in the sequence $x(1), a(2), \hdots, x(t), a(t)$ for set $J \in \p$, and by $N_t(J, a_J)$ the number of time action $a_J$ was chosen at time $t$. Then, conditioned on $\bigcup_{J \in \p} \G_J$, for every $t \geq 0$
    \begin{align*}
        \ell_t 
        &= 
        \sum_{i=1}^n \brk[s]*{\brk[a]*{x(t), \theta^*_{i,a^*_i(t)}} - \brk[a]*{x(t), \theta^*_{i,a_i(t)}}} \\
        &=
        \sum_{J \in \p} \sum_{i \in J} \brk[s]*{\brk[a]*{x(t), \theta^*_{i,a^*_i(t)}} - \brk[a]*{x(t), \theta^*_{i,a_i(t)}}} \\
        &=
        \sum_{J \in \p}
        \brk[s]*{\brk[a]*{x(t), \theta^*_{a_J^*(t)}} - \brk[a]*{x(t), \theta^*_{a_J(t)}}} \\
        &\leq
        \sum_{J \in \p}
        2\sqrt{\beta_J(t)}
        \norm{x(t)}_{V_{J, a_J}^{-1}}
    \end{align*}
    Next, notice that $\ell_t \leq 2$ since $\sum_{i\in J} \brk[a]*{x(t), \theta^*_{i,a}} \in [-1, 1]$. Therefore,
    \begin{align*}
        \ell_t 
        \leq
        \sum_{J \in \p}
        2\sqrt{\beta_J(t)}
        \min\brk[c]*{\norm{x(t)}_{V_{J, a_J}^{-1}}, 1}.
    \end{align*}
    Combining the above we get that conditioned on $\bigcup_{J \in \p} \G_J$, for every $t \geq 0$
    \begin{align*}
        R(T)
        &\leq
        \sqrt{T\sum_{t=0}^T \ell_t^2} \\
        &\leq
        2\sqrt{T\sum_{J \in \p} \beta_J(t) \sum_{t=0}^T \min\brk[c]*{\norm{x(t)}^2_{V_{J, a_J}^{-1}}, 1}} \\
        &=
        2\sqrt{T\sum_{J \in \p} \beta_J(t) \sum_{a_J}\sum_{k=1}^{N_T(J, a_J)} \min\brk[c]*{\norm{x(I_k(J, a_J))}^2_{V_{J, a_J}(I_k(J, a_J))^{-1}}, 1}} \\
        &\leq
        2\sqrt{T}
        \brk*{
        \lambda^{1/2}nS_\theta + R_{\text{max}} \sqrt{d\log\brk*{\frac{\abs{\p} K^n (1 + tS_x) / \lambda}{\delta}}}} 
        \sqrt{
        \sum_{J \in \p}
        \sum_{a_J}\sum_{k=1}^{N_T(J, a_J)} \min\brk[c]*{\norm{x(I_k(J, a_J))}^2_{V_{J, a_J}(I_k(J, a_J))^{-1}}, 1}} \\
        &\overset{(1)}{\leq}
        2\sqrt{T}
        \brk*{
        \lambda^{1/2}nS_\theta + R_{\text{max}} \sqrt{d\log\brk*{\frac{\abs{\p} K^n (1 + tS_x) / \lambda}{\delta}}}}
        \sqrt{
        d
        \sum_{J \in \p}
        \sum_{a_J}
        \log\brk*{\lambda + \frac{N_T(J, a_J) S_x^2}{d}}
        } \\
        &\overset{(2)}{\leq}
        2\sqrt{T}
        \brk*{
        \lambda^{1/2}nS_\theta + R_{\text{max}} \sqrt{d\log\brk*{\frac{\abs{\p} K^n (1 + tS_x) / \lambda}{\delta}}}}
        \sqrt{
        d
        \sum_{J \in \p}
        K^{\abs{J}}
        \log\brk*{\lambda + \frac{T S_x^2}{K^{\abs{J}}d}}
        } \\
        &\leq
        2\sqrt{T}
        \brk*{
        \lambda^{1/2}nS_\theta + R_{\text{max}} \sqrt{d\log\brk*{\frac{\abs{\p} K^n (1 + tS_x) / \lambda}{\delta}}}}
        \sqrt{
        d
        \log\brk*{\lambda + \frac{T S_x^2}{Kd}}
        \sum_{J \in \p}
        K^{\abs{J}}
        }
    \end{align*}
    where in $(1)$ we used Lemma~1 of \citet{abbasi2011improved}, and in $(2)$ Jensen's inequality and the fact that $\sum_{a_J} N_T(J, a_J) = T$, i.e.,
    \begin{align*}
        \sum_{a_J}
        \log\brk*{\lambda + \frac{N_T(J, a_J) S_x^2}{d}}
        &=
        K^{\abs{J}}
        \sum_{a_J}
        \frac{1}{K^{\abs{J}}}
        \log\brk*{\lambda + \frac{N_T(J, a_J) S_x^2}{d}} \\
        &\leq
        K^{\abs{J}}
        \log\brk*{\lambda + \frac{\sum_{a_J} \frac{1}{K^{\abs{J}}}N_T(J, a_J) S_x^2}{d}} \\
        &=
        K^{\abs{J}}
        \log\brk*{\lambda + \frac{T S_x^2}{K^{\abs{J}}d}}.
    \end{align*}
    This completes the proof.
\end{proof}

\begin{lemma}
\label{lemma: confidence ellipsoid}
    Denote $\theta^*_{a_J} = \sum_{i \in J}\theta^*_{i, a_i}$ and let $\hat{\theta}_{a_J}(t) = V_{J, a_J}(t)^{-1}Y_J(t)$ be the least squares estimator at time step $t$, where $
         V_{J, a_J}(t)
         =
         \lambda I
         +
         \sum_{k=1}^{t-1}
         \mathbf{1}_{\brk[c]*{a_J(k) = a_J}}x(k)x(k)^T$.
    Let ${\sqrt{\beta_J(t, \delta)} = \lambda^{1/2}\abs{J}S_\theta + R_{\text{max}} \sqrt{d\log\brk*{\frac{\abs{\p} K^{\abs{J}} (1 + tS_x) / \lambda}{\delta}}}}$ and
    \begin{align*}
        \mathcal{C}_{t,a_J}
        =
        \brk[c]*{\theta \in \R^d : \norm{\theta_{a_J} - \hat{\theta}_{a_J}(t)}_{V_{J, a_J}} \leq \sqrt{\beta_J(t)}}.
    \end{align*}
    Then, for all $t \geq 0, a_J \in \bigtimes_{i \in J} \A_i$, with probability at least $1-\frac{\delta}{\abs{\p}}$,
    $
        \theta^*_{a_J} \in \mathcal{C}_{t,a_J}.
    $
\end{lemma}
\begin{proof}
    Employing Theorem~2 of \citet{abbasi2011improved} with
    \begin{align*}
        \norm{\theta^*_{a_J}}_2 = \norm{\sum_{i \in J}\theta^*_{i, a_i}}_2 \leq \abs{J}S_\theta
    \end{align*}
    and taking the Union bound over all $a_J \in \bigtimes_{i \in J} \A_i$ yields that with probability at least $1-\delta$, 
    \begin{align*}
        \norm{\theta_{a_J} - \hat{\theta}_{a_J}(t)}_{V_{J, a_J}}
        \leq
        \lambda^{1/2}\abs{J}S_\theta + R_{\text{max}} \sqrt{d\log\brk*{\frac{K^{\abs{J}} (1 + tS_x) / \lambda}{\delta}}}.
    \end{align*}
    Using $\tilde{\delta} = \frac{\delta}{\abs{\p}}$ completes the proof.
\end{proof}

\end{document}